\documentclass{article} 
\usepackage[accepted]{icml2024}

\usepackage{microtype}
\usepackage{graphicx}
\usepackage{booktabs} 

\usepackage{hyperref}

\usepackage{url}
\usepackage{amsmath,amsfonts,bm}

\usepackage[final]{pdfpages}
\usepackage{amsthm}

\usepackage{xspace, multirow}
\usepackage{flushend}
\usepackage[font=bf]{caption}
\usepackage[ruled, vlined, linesnumbered,algo2e]{algorithm2e}
\usepackage{colortbl}
\usepackage{bbold}
\usepackage{enumitem}

\usepackage{tikz}
\usepackage{pgfplots}
\usetikzlibrary{patterns}
\pgfplotsset{compat=1.16}
\usepackage[captionskip=-4pt]{subfig}
\usepackage{wrapfig}

\usepackage{tablefootnote}

\newcommand{\ie}{{i.e.,}\xspace}
\newcommand{\eg}{{e.g.,}\xspace}
\newcommand{\spara}[1]{\vspace{1mm}\noindent\textbf{#1.}}
\newcommand{\savespace}[1]{\ignorespaces}

\newcommand{\SGC}{SGC\xspace}
\newcommand{\SIGN}{SIGN\xspace}
\newcommand{\ChebNet}{ChebNet\xspace}
\newcommand{\ChebNetII}{ChebNetII\xspace}
\newcommand{\GPRGNN}{GPR-GNN\xspace}
\newcommand{\BernNet}{BernNet\xspace}
\newcommand{\JacobiConv}{JacobiConv\xspace}
\newcommand{\ASGC}{ASGC\xspace}
\newcommand{\OptBasisGNN}{OptBasisGNN\xspace}
\newcommand{\newbasis}{UniBasis\xspace}
\newcommand{\ours}{UniFilter\xspace}

\newcommand{\GCN}{GCN\xspace}
\newcommand{\GCNII}{GCNII\xspace}
\newcommand{\GAT}{GAT\xspace}
\newcommand{\MixHop}{MixHop\xspace}
\newcommand{\EvenNet}{EvenNet\xspace}
\newcommand{\HGCN}{H$_2$GCN\xspace}
\newcommand{\LINKX}{LINKX\xspace}
\newcommand{\ACM}{ACM-GCN\xspace}
\newcommand{\GloGNN}{GloGNN++\xspace}
\newcommand{\WRGAT}{WRGAT\xspace}
\newcommand{\Specformer}{Specformer\xspace}
\newcommand{\OrderedGNN}{Ordered GNN\xspace}

\newcommand{\HomFilter}{HomFilter\xspace}
\newcommand{\HetFilter}{HetFilter\xspace}
\newcommand{\OrtFilter}{OrtFilter\xspace}

\newcommand{\A}{\mathbf{A}\xspace}

\newcommand{\D}{\mathbf{D}\xspace}
\newcommand{\G}{\mathbf{G}\xspace}

\newcommand{\I}{\mathbf{I}\xspace}
\renewcommand{\P}{\mathbf{P}\xspace}
\newcommand{\U}{\mathbf{U}\xspace}

\newcommand{\X}{\mathbf{X}\xspace}
\newcommand{\Y}{\mathbf{Y}\xspace}
\newcommand{\Z}{\mathbf{Z}\xspace}
\newcommand{\bLambda}{\mathbf{\Lambda}\xspace}

\renewcommand{\L}{\mathcal{L}\xspace}

\newcommand{\TD}{\tilde{\D}\xspace}
\newcommand{\TA}{\tilde{\A}\xspace}

\newcommand{\Tlambda}{\tilde{\lambda}\xspace}

\newcommand{\h}{\mathbf{h}\xspace}

\newcommand{\s}{\mathbf{s}\xspace}

\newcommand{\w}{\mathbf{w}\xspace}
\newcommand{\x}{\mathbf{x}\xspace}
\newcommand{\y}{\mathbf{y}\xspace}
\renewcommand{\v}{\mathbf{v}\xspace}
\renewcommand{\u}{\mathbf{u}\xspace}
\newcommand{\z}{\mathbf{z}\xspace}

\newcommand{\C}{\mathcal{C}\xspace}
\newcommand{\R}{\mathbb{R}\xspace}
\newcommand{\N}{\mathcal{N}\xspace}
\newcommand{\F}{\mathcal{F}\xspace}

\newcommand{\V}{\mathcal{V}\xspace}
\newcommand{\E}{\mathcal{E}\xspace}

\newcommand{\g}{\ensuremath{\mathrm{g}}}

\theoremstyle{plain}
\newtheorem{theorem}{Theorem}[section]
\newtheorem{proposition}[theorem]{Proposition}
\newtheorem{lemma}[theorem]{Lemma}

\theoremstyle{definition}
\newtheorem{definition}[theorem]{Definition}

\theoremstyle{remark}

\newcommand{\eat}[1]{}
\newcommand{\revise}[1]{{#1}}
\newcommand{\del}[1]{}

\newenvironment{customlegend}[1][]{%
    \begingroup
    \csname pgfplots@init@cleared@structures\endcsname
    \pgfplotsset{#1}%
}{%
    \csname pgfplots@createlegend\endcsname
    \endgroup
}%

\def\addlegendimage{\csname pgfplots@addlegendimage\endcsname}

\icmltitlerunning{How Universal Polynomial Bases Enhance Spectral Graph Neural Networks}

\begin{document}

\twocolumn[
\icmltitle{How Universal Polynomial Bases Enhance Spectral Graph Neural Networks: Heterophily, Over-smoothing, and Over-squashing} 

\icmlsetsymbol{equal}{*}

\begin{icmlauthorlist}
\icmlauthor{Keke Huang}{nus}
\icmlauthor{Yu Guang Wang}{sjtu1,sjtu2,sjtu3}
\icmlauthor{Ming Li}{zj1,zj2}
\icmlauthor{Pietro Li\`{o}}{cam}
\end{icmlauthorlist}

\icmlaffiliation{nus}{School of Computing, National University of Singapore, Singapore}
\icmlaffiliation{sjtu1}{Institute of Natural Sciences, School of Mathematical Sciences, Zhangjiang Institute for Advanced Study, Shanghai Jiao Tong University, Shanghai, China}
\icmlaffiliation{sjtu2}{Shanghai AI Laboratory, Shanghai, China}
\icmlaffiliation{sjtu3}{School of Mathematics and Statistics, University of New South Wales, Sydney, Australia}
\icmlaffiliation{zj1}{Zhejiang Institute of Optoelectronics, Jinhua, China}
\icmlaffiliation{zj2}{Zhejiang Key Laboratory of Intelligent Education Technology and Application, Zhejiang Normal University, Jinhua, China}
\icmlaffiliation{cam}{Department of Computer Science and Technology, Cambridge University, Cambridge, UK}

\icmlcorrespondingauthor{Keke Huang}{kkhuang@nus.edu.sg}
\icmlcorrespondingauthor{Ming Li}{mingli@zjnu.edu.cn}

\icmlkeywords{Graph Neural Networks, Polynomial Graph Filters, Heterophily}

\vskip 0.3in
]

%

\printAffiliationsAndNotice{}  

\begin{abstract}

Spectral Graph Neural Networks (GNNs), alternatively known as {\em graph filters}, have gained increasing prevalence for heterophily graphs. Optimal graph filters rely on Laplacian eigendecomposition for Fourier transform. In an attempt to avert prohibitive computations, numerous polynomial filters have been proposed. However, polynomials in the majority of these filters are {\em predefined} and remain {\em fixed} across different graphs, failing to accommodate the varying degrees of heterophily. Addressing this gap, we demystify the intrinsic correlation between the spectral property of desired polynomial bases and the heterophily degrees via thorough theoretical analyses. Subsequently, we develop a novel adaptive heterophily basis wherein the basis vectors mutually form angles reflecting the heterophily degree of the graph. We integrate this heterophily basis with the homophily basis to construct a universal polynomial basis {\em \newbasis}, which devises a polynomial filter-based graph neural network -- {\em \ours}. It optimizes the convolution and propagation in GNN, thus effectively limiting over-smoothing and alleviating over-squashing. Our extensive experiments, conducted on datasets with a diverse range of heterophily, support the superiority of \newbasis in the universality but also its proficiency in graph explanation.  
\end{abstract}

\begin{sloppy}
\section{Introduction}\label{sec:intro} 

Spectral Graph Neural Networks (GNNs)~\citep{KipfW17}, also referred to as {\em graph filters}, have been extensively investigated in recent years due to their superior performance in handling heterophily graphs~\cite{abs-2302-05631}. Optimal graph filters conduct Laplacian eigendecomposition for Fourier transform. To bypass computation complexity, existing graph filters leverage various polynomials to approximate the desired filters for graphs with varying heterophily degrees. For example, \ChebNet~\citep{DefferrardBV16} employs truncated Chebyshev polynomials~\citep{mason2002chebyshev, hammond2011wavelets} and accomplishes localized spectral filtering. \BernNet~\citep{he2021bernnet} utilizes Bernstein polynomials~\citep{Farouki12} to acquire better controllability and interpretability. Later, \citet{WangZ22} propose \JacobiConv by exploiting Jacobi polynomial bases~\citep{askey1974positive} with improved generality.

However, existing polynomial filters overlook the diverse heterophily degrees of underlying graphs when implementing polynomial bases. This oversight of the graph homophily characteristic in the development of polynomial bases leads to suboptimal performance on real-world graphs, as demonstrated in our experiments (Sections~\ref{sec:classresults} and~\ref{sec:ablation}). Meanwhile, Theorem~\ref{thm:frequencyratio} proves that frequencies of signals filtered by optimal graph filters are proportional to the heterophily degrees. Therefore, ideal polynomial bases are obligated to provide adaptability to the wide range of heterophily degrees. Consequently, a natural question arises: {\bf how can we devise a universal polynomial basis that encapsulates the diverse graph heterophily degrees?} 

To address the issue, we initially determine the relationship between the degree of heterophily and the frequency of optimally filtered signals
(Theorem~\ref{thm:frequencyratio}).
Following this, we investigate how the distribution of polynomial bases within the Euclidean space affects the basis spectrum on regular graphs, as outlined in Theorem~\ref{thm:pivot}. These pivotal insights pave the way for creating an innovative adaptive heterophily basis, specifically designed to match the heterophily degrees in graphs. Ultimately, we combine the heterophily and homophily bases to establish a comprehensive universal basis, termed {\em\newbasis}. Utilizing \newbasis, we develop a graph neural network using a general polynomial filter, named {\em \ours}, which is adaptable to various graph structures.
 
Many existing spectral and spatial GNNs usually suffer from over-smoothing~\cite{LiHW18, NguyenHNHON23,abs-2303-10993} and over-squashing~\cite{AlonY21, ToppingGC0B22, GiovanniGBLLB23}. These scenarios occur when node features go consensus and exponential information is squeezed into fixed-sized vectors in long-range propagation. However, our approach, \ours, effectively tackles these two pivotal challenges by employing a convolutional matrix performing both information propagation and rotation concurrently, thereby effectively rewiring the graph structures. 
For a comprehensive evaluation, we compare \ours with $20$ baselines across both real-world datasets and synthetic datasets with a wide range of heterophily degrees. The consistently outstanding performance of \ours across these datasets validates its efficacy and wide applicability. Moreover, we showcase the spectrum distribution of graph signals as captured by \newbasis across various datasets in Section~\ref{sec:spectrum}. These findings clearly underscore \newbasis's potent potential as a groundbreaking tool for graph explanation, thereby deepening our comprehension of graph structures.

In summary, 1) we reveal that polynomials of desired polynomial filters are meant to align with degrees of graph heterophily; 2) we design a universal polynomial basis \newbasis tailored for different graph heterophily degrees and devise a general graph filter \ours; 3) we prove in theory that \ours effectively averts over-smoothing while alleviating over-squashing; 4) we evaluate \ours on both real-world and synthetic datasets against $20$ baselines. The exceptional performance of \ours not only attests to the effectiveness and broad applicability of \newbasis but also highlights its significant potential in enhancing graph explanation. The code of \ours is accessed at \url{https://github.com/kkhuang81/UniFilter}.

\section{Preliminaries}\label{sec:preliminary}

\spara{Notations and Definitions} We represent matrices, vectors, and sets with bold uppercase letters (\eg $\A$), bold lowercase letters (\eg $\x$), and calligraphic fonts (\eg $\N$), respectively. The $i$-th row (resp.\ column) of matrix $\A$ is represented by $\A[i,\cdot]$ (resp.\ $\A[\cdot, i]$). We denote $[n]=\{1,2,\cdots, n\}$.  

Let $\G=(\V, \E)$ be an undirected and connected graph with node set $|\V|=n$ and edge set $|\E|=m$. Let $\X \in \R^{n\times d}$ be the $d$-dimension feature matrix. For ease of exposition, we use node notation $u\in \V$ to denote its index, \ie $\X_u=\X[u,\cdot]$. Let $\Y\in \mathbb{N}^{n\times |\C|}$ be the one-hot label matrix, \ie $\Y[u,i]=1$ if node $u$ belongs to class $\C_i$ for $i\in [|\C|]$, where $\C$ is the set of node labels. The neighbor set of node $u\in \V$ is denoted as $\N_u$ with degree $d_u=|\N_u|$. The adjacency matrix of $\G$ is denoted as $\A \in \R^{n\times n}$ that $\A[u,v]=1$ if edge $\langle u, v \rangle \in \E$; otherwise $\A[u,v]=0$. $\D \in \R^{n\times n}$ is the diagonal degree matrix of $\G$ with $\D[u,u]=d_u$. Let $\L$ and $\hat{\L}$ be the normalized Laplacian matrix of graph $\G$ without and with self-loops respectively, defined as $\L=\I-\D^{-\frac{1}{2}}\A\D^{-\frac{1}{2}}$ and $\hat{\L}=\I-\TD^{-\frac{1}{2}}\TA\TD^{-\frac{1}{2}}$ where $\I$ is the identity matrix and $\TD=\D+\I$ and $\TA=\A+\I$. Frequently used notations are summarized in Table~\ref{tbl:notations} in Appendix~\ref{app:notations}.

\spara{Spectral graph filters} In general, the eigendecomposition of the Laplacian matrix is denoted as $\L=\U \bLambda \U^\top$, where $\U$ is the eigenvector matrix and $\bLambda=\mathrm{diag}[\lambda_1, \cdots, \lambda_n]$ is the diagonal matrix of eigenvalues. Eigenvalues $\lambda_i$ for $i \in [n]$ mark the {\em frequency} and the eigenvalue set $\{\lambda_1, \cdots, \lambda_n\}$ is the {\em graph spectrum}. Without loss of generality, we assume $0=\lambda_1 \le \lambda_2 \le \cdots \le \lambda_n \le 2$. When applying a spectral graph filter on graph signal $\x \in \R^n$, the process involves the following steps. First, the graph Fourier operator $\F(\x)=\U^\top \x$ projects the graph signal $\x$ into the spectral domain. Next, a spectral filtering function $\g_\w(\cdot)$ parameterized by $\w\in \R^n$ is applied on the derived spectrum. Last, the filtered signal is transformed back via the inverse graph Fourier transform operator $\F^{-1}(\x)=\U \x$. The process is 
\begin{align}
&\F^{-1}(\F(\g_\w) \odot  \F(\x))\notag=\U\g_\w(\bLambda)\U^\top \x\notag\\ 
&=\U\ \mathrm{diag}(\g_\w(\lambda_1), \cdots, \g_\w(\lambda_n))\U^\top \x, \label{eqn:Fourier} 
\end{align}
where $\odot$ is the Hadamard product. 

Spectral graph filters enhance signals in specific spectrums and suppress the rest parts according to objective functions. For node classification, homophily graphs are prone to contain low-frequency signals whilst heterophily graphs likely own high-frequency signals. 
To quantify the heterophily degrees of graphs, numerous homophily metrics have been introduced, \eg\ {\em edge homophily}~\citep{ZhuYZHAK20}, {\em node homophily}~\citep{PeiWCLY20}, {\em class homophily}~\citep{lim2021large}, and a recent {\em adjusted homophily}~\citep{platonov2022}. By following the literature on spectral graph filters~\citep{ZhuYZHAK20, LeiWLDW22}, we adopt edge homophily in this work. 

\begin{definition}[Homophily Ratio $h$]~\label{def:homo}
Given a graph $\G=(\V,\E)$ and its label matrix $\Y$, the homophily ratio $h$ of $\G$ is the fraction of edges with two end nodes from the same class, \ie $h=\textstyle\frac{|\{\langle u,v \rangle \in \E \colon \y_u=\y_v\} |}{|\E|}$.
\end{definition} 

\spara{Discussion on homophily metrics} \citet{lim2021large,platonov2022} point out that homophily metrics quantify the tendency of nodes to connect other nodes from the same classes and measure the deviation of the label distribution from a null model, denoted as {\em constant baseline} property. In this regard, it is supposed to eliminate the homophily degrees stemming from random edge connections. Yet in our paper, what matters in terms of message-passing GNNs is whether edges connect nodes from the same classes. Generally, the more two endpoints of edges are from the same classes in graphs, the more beneficial the message passing for information aggregation. Therefore, we adopt edge homophily as the homophily metric.

In addition to the homophily metrics for {\em categorical} node labels, the similarity of {\em numerical} node signals can also be measured via {\em Dirichlet Enenrgy}~\citep{ZhouHZCLCH21, KarhadkarBM23} (Detailed discussion in Section~\ref{sec:theory}). In terms of spectral perspective, we propose {\em spectral signal frequency}, a metric customized for node signals $\x\in \R^n$.

\begin{definition}[Spectral Signal Frequency $f$]~\label{def:frequency}
Consider a graph $\G=(\V,\E)$ with $n$ nodes and Laplacian matrix $\L$. Given a normalized feature signal $\x\in \R^n$, the spectral signal frequency $f(\x)$ on $\G$ is defined as $f(\x)=\tfrac{\x^\top\L\x}{2}$.
\end{definition} 
By nature of the Laplacian matrix, spectral signal frequency $f(x)$ quantifies the discrepancy of signal $\x$ on graph $\G$. Formally, the spectral signal frequency $f(x)$ holds that
\begin{proposition}\label{pro:frequencybound}
For any normalized feature signal $\x \in \R^{n}$ on graph $\G$, $f(\x)\in [0,1]$.  
\end{proposition}

\section{Polynomial Graph Filters}\label{sec:revisting}

Optimal graph filters perform eigendecomposition on the Laplacian matrix with a computation cost of $O(n^3)$. To bypass the significant computation overhead, substantial polynomial graph filters~\citep{ChienP0M21, WangZ22, HeWW22, GuoW23,huang2024Optimizing} have been proposed to approximate optimal graph filters by leveraging different polynomials (see Table~\ref{tbl:pgf} in Appendex~\ref{app:notations}). By identifying an appropriate propagation matrix $\P$, those polynomial filters on graph signal $\x\in \R^n$ is equally expressed as 
\begin{equation}\label{eqn:gsf}
\z=\textstyle\sum^K_{k=0}\w_k \P^k\cdot \x,   
\end{equation}
where $K$ is the propagation hops, $\w\in \R^{K+1}$ is the learnable weight vector, and $\z\in \R^n$ is the final representation. For example, \BernNet~\cite{he2021bernnet} utilizes Bernstein polynomial as $\z=\textstyle\sum^K_{k=0}\frac{\w^\prime_k}{2^K}\binom{K}{k}(2\I-\L)^{K-k}\L^k \x$. By setting $\P=\I-\frac{\L}{2}$ and reorganizing the equation, we derive an equivalent formulation as $\z=\textstyle\sum^K_{k=0}\w_k \left(\I-\frac{\L}{2}\right)^k \x$ where \[\w_k=\sum^k_{i=0} \w^\prime_{k-i}\binom{K}{K-i}\binom{K-i}{k-i}(-1)^{k-i}\] works as the new learnable parameter. 

In Equation~\eqref{eqn:gsf}, the vectors $\P^k\x$ for $k\in \{0,1,\cdots, K\}$ collectively constitute a signal basis $\{\P^0\x, \P^1\x, \cdots, \P^K\x\}$. Spectral graph filters aim to generate node representations within graphs with various heterophily degrees by learning a weighted combination of the signal basis. From the spectral perspective, they essentially derive signals with the desired frequency from the spectrum $\{f(\P^0\x), f(\P^1\x), \cdots, f(\P^K\x)\}$, ensuring alignment with label signals $\Y$. According to Definition~\ref{def:homo} and~\ref{def:frequency}, the homophily ratio $h$ and the frequency of label signals attempt to epitomize the inconsistency of label information distributed on graphs. This fact manifests the inherent correlation between the filtered signal $\textstyle\sum^K_{k=0}\w_k\P^k\x$ and $h$. Therefore, we propose a theorem to formally depict the correlation.

\begin{theorem}\label{thm:frequencyratio}
Given a connected graph $\G=(\V,\E)$ with homophily ratio $h$, consider an optimal polynomial filter $\mathrm{F(\w)}=\textstyle\sum^K_{k=0}\w_k \P^k$ with propagation matrix $\P$ and weights $\w \in \R^{K+1}$ for node classification. Given a feature signal $\x\in \R^n$, the spectral frequency $f(\textstyle\sum^K_{k=0}\w_k\P^k\x)$ is proportional to $1\!-\!h$. 
\end{theorem}

Theorem~\ref{thm:frequencyratio} clarifies that ideal signal bases are obligated to incorporate the heterophily properties of graphs. This implies that high-frequency signals can be more desirable on heterophily graphs. However, the majority of existing polynomial filters ignore the graph homophily ratios while relying on predefined polynomials instead, thereby yielding suboptimal performance.

\section{Universal Polynomial Basis for Graph Filters}\label{sec:unibasis}

\begin{figure}
\centering
\captionsetup[subfigure]{margin={0.2cm,0.0cm}}
\subfloat[{Trend of homophily basis}]{\includegraphics[height=0.46\columnwidth]{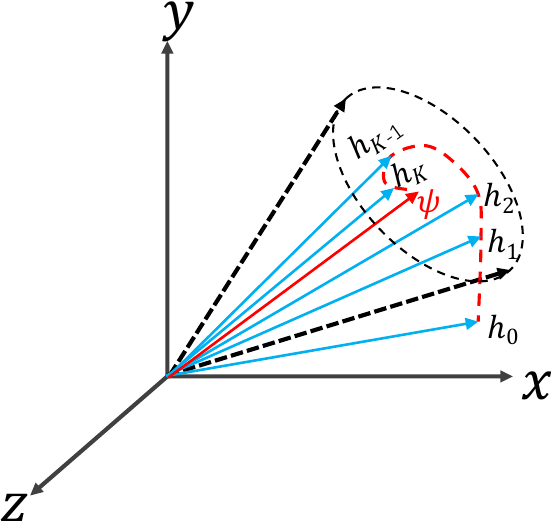}\label{subfig:homo}}\hspace{1mm}
\subfloat[{Construction of heterophily basis}]{\includegraphics[height=0.46\columnwidth]{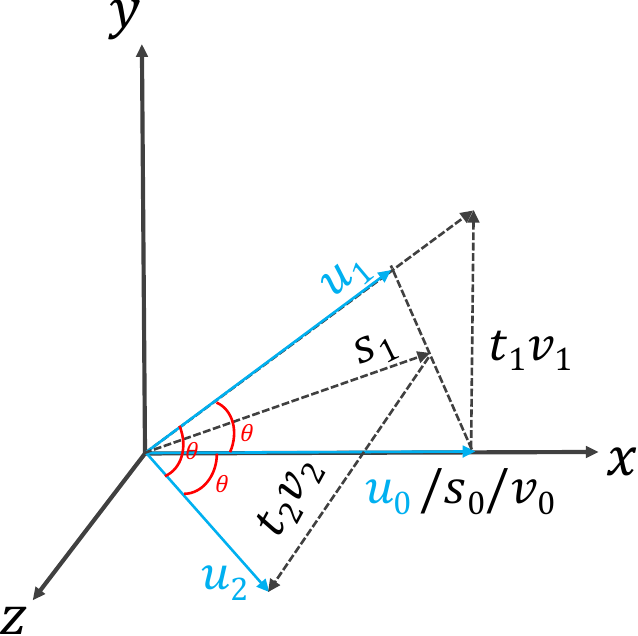}\label{subfig:hetero}}
\caption{Left: Converging trend of homophily bases. Right: Construction procedure of heterophily bases.}\label{fig:homoheterbasis}
\end{figure}

In this section, we develop a novel universal polynomial basis, denoted as {\em \newbasis}, and subsequently we present a general polynomial filter, referred to as {\em \ours}, built upon \newbasis. Specifically, the construction of \newbasis involves the integration of two foundational components: the traditional homophily basis $\{\x, \P\x, \cdots, \P^K\x\}$ (Section~\ref{sec:homobasis}), and an innovative adaptive heterophily basis $\{\u_0,\u_1,\cdots,\u_K\}$ (Section~\ref{sec:heterobasis}), governed by a hyperparameter $\tau\in[0,1]$. This integration, formulated as $\tau\P^k \x+(1-\tau)\u_k$, enables \newbasis the adaptability for a variety of heterophily graphs. Consequently, the polynomial filter \ours is crafted as 
\begin{equation}\label{eqn:generalbasis}
\z=\textstyle\sum^K_{k=0}\w_k (\tau\P^k \x+(1-\tau)\u_k), 
\end{equation}
where vector $\w \in \R^{K+1}$ is the learnable weight. Subsequently, $\z$ is fed into Multilayer Perceptron (MLP) network for weight training, \ie \[\mathrm{\y}=\mathrm{Softmax}(\mathrm{MLP}(\sum^K_{k=0}\mathbf{w}_k(\tau\mathbf{P}^k\mathbf{x}+(1-\tau)\mathbf{u}_k))).\] In the following sections, we first establish the property of homophily basis and then elaborate on the heterophily basis design. Ultimately, we introduce the graph explanation capability of \newbasis.

\subsection{Theoretical Analysis of Homophily Basis}\label{sec:homobasis}

Conventional GNN models~\citep{HamiltonYL17, KlicperaBG19,YangSZNWCW23,huang2023node} employ homophily as a strong inductive bias~\citep{lim2021large}. In general, graph signal $\x$ is propagated to $K$-hop neighbors via propagation matrix $\P=\I-\L$ for information aggregation, yielding a $(K+1)$-length {\em homophily basis} $\{\x, \P\x,\cdots,\P^K\x\}$. We formulate a theorem to explore the property of the homophily basis, shedding light on its potential to accommodate homophily graphs.
\begin{theorem}\label{thm:homo}
Given a propagation matrix $\P$ and graph signal $\x$, consider an infinite homophily basis $\{\x, \P\x, \cdots, \P^k\x, \P^{k+1}\x, \cdots\}$. There exists an integer $\eta \in \mathbb{N}$ such that when the exponent $k\ge \eta$ and increases, the angle $\arccos\left(\tfrac{\P^k\x \cdot \P^{k+1}\x}{\|\P^k\x\| \|\P^{k+1}\x\|}\right)$ is progressively smaller and asymptotically approaches $0$.
\end{theorem} 

The homophily basis exhibits {\em growing similarity} and {\em asymptotic convergence} to capture homophily signals. This phenomenon, however, leads to the {\em over-smoothing issue}. For better visualization, Figure~\ref{subfig:homo} illustrates the converging trend of a homophily basis in Euclidean space.

\subsection{Adaptive Heterophily Basis}\label{sec:heterobasis}

As proved in Theorem~\ref{thm:frequencyratio}, the desired signal bases are expected to adapt to homophily ratios. This raises a pertinent question: {\em how can we incorporate homophily ratios in a sensible manner when designing signal bases?} To answer this question, we initially explore the correlation between the basis distribution and its frequency on regular graphs for simplicity, deriving insights for general cases.

\begin{theorem}\label{thm:pivot}
Consider a regular graph $\G$, a random basis signal $\x \in \R^n$, and a normalized all-ones vector $\phi\in \R^n$ with $f(\phi)=0$. Suppose $\theta:=\arccos( \phi \cdot \x )$ denotes the angle formed by $\x$ and $\phi$. It holds that the expectation of spectral signal frequency $\mathbb{E}_{\G\sim \mathcal{G}}[f(\x)]$ over the randomness of $\G$ is monotonically increasing with $\theta$ for $\theta \in [0,\tfrac{\pi}{2})$.
\end{theorem}

\spara{Heterophily basis construction} Heterophily bases aim to align the spectrum property to the homophily ratios of underlying graphs. Theorem~\ref{thm:pivot} establishes that signal vectors exhibiting a greater angular separation from the $0$-frequency vector $\phi$ possess a higher expected frequency on regular graphs. When attempting to extend to general graphs without involving the $0$-frequency vector, we may consider the angular offsets (relative position) among the basis vectors themselves for the desired spectrum property. 

Theorem~\ref{thm:homo} discloses the growing similarity and asymptotic convergence phenomenon within the homophily basis, \ie angles among homophily basis approaching zero. To mitigate this over-smoothing issue, we can intuitively scatter and fix an angle of $\theta \in [0,\tfrac{\pi}{2}]$ among all pairs of basis vectors, as inspired by the insight from Theorem~\ref{thm:pivot}. Then, a natural question arises: {\em how do we determine an appropriate value for $\theta$?}

Specifically, Theorem~\ref{thm:frequencyratio} proves the spectral frequency of ideal signals proportional to $1-h$. Meanwhile, Theorem~\ref{thm:pivot} hints at the monotonic relation between the spectral frequency and angular offsets among basis vectors. By integrating these insights of linear proportion and monotonic relation, we empirically set the $\theta:=\frac{\pi}{2}(1-h)$. Consequently, we derive a signal basis adept at encapsulating the varying heterophily degrees of graphs, which is formally denoted as the {\em heterophily basis}.

The procedure of constructing a $(K\!+\!1)$-length of heterophily basis is outlined in Algorithm~\ref{alg:basis} and illustrated in Figure~\ref{subfig:hetero}. To start with, we normalize the input signal $\x$ as the initial signal $\u_0$ and set $\theta:=\tfrac{(1-\hat{h})\pi}{2}$ where $\hat{h}$ is the estimation of $h$. To manipulate the formed angles between signal vectors, we forge an orthonormal basis, denoted as $\{\v_0, \v_1, \cdots, \v_K\}$ where $\v_0$ is initialized as $\u_0$. In particular, at the $k$-th iteration for $k\in[1, K]$, we set $\v_k := \P \v_{k-1}$. Subsequently, $\v_k$ is calculated as $\v_k:= \v_k-  (\v^\top_k \v_{k-1})\v_{k-1} - (\v^\top_k \v_{k-2}) \v_{k-2}$ as per the {\em three-term recurrence theorem}~\citep{gautschi2004orthogonal,liesen2013krylov,GuoW23}. Signal vector $\u_k$ is set as $\u_k:=\tfrac{\s_{k-1}}{k}$ where $\s_{k-1}:=\sum^{k-1}_{i=0}\u_i$. Subsequently, $\u_k$ is updated as $\u_k := \tfrac{\u_k+t_k\v_k}{\|\u_k+t_k\v_k\|}$ where $t_k$ is
\begin{equation}\label{eqn:factor}
t_k=\sqrt{\left(\tfrac{\s_{k-1}^\top\u_{k-1}}{k\cos\theta}\right)^2-\tfrac{(k-1)\cos\theta+1}{k}}.    
\end{equation}
The final vector set $\{\u_0,\u_1,\cdots,\u_K\}$ is returned as the heterophily basis. The desired property of the heterophily basis is proved in the following theorem. Detailed proofs are presented in Appendix~\ref{app:proofs}.

\begin{algorithm}
\caption{Heterophily Basis}
\label{alg:basis}
\KwIn{Graph $\G$, propagation matrix $\P$, input feature signal $\x$, hop $K$, \revise{estimated homophily ratio $\hat{h}$}}
\KwOut{Heterophily basis $\{\u_0,\u_1,\cdots,\u_K\}$}
$\u_0\gets \tfrac{\x}{\|\x\|}$, $\v_0 \gets \u_0$, $\v_{-1} \gets \mathbf{0}$, $\s_0\gets \u_0$, \revise{$\theta \gets \tfrac{(1-\hat{h})\pi}{2}$}\;
\For{$k \gets 1$ \KwTo $K$} 
{
    $\v_k\gets \P\v_{k-1}$\;
    
    $\v_k \gets \v_k-  (\v^\top_k \v_{k-1})\v_{k-1} -  (\v^\top_k \v_{k-2}) \v_{k-2}$\;
    
    $\v_k \gets \tfrac{\v_k}{\|\v_k\|}$, $\u_k\gets \textstyle\tfrac{\s_{k-1}}{k}$\;
    
    $t_k$ is calculated as in Equation~\eqref{eqn:factor}\;
    
    $\u_k \gets \tfrac{\u_k+t_k\v_k}{\|\u_k+t_k\v_k\|}$, $\s_k \gets \s_{k-1}+\u_k$\;
}
\Return $\{\u_0,\u_1,\cdots,\u_K\}$\;
\end{algorithm}

\begin{theorem}\label{thm:heteroproperty}
Consider a heterophily basis $\{\u_0,\u_1, \cdots \u_K\}$ constructed from Algorithm~\ref{alg:basis} for graphs with homophily ratio $h$. It holds that $\forall i,j\in \{0,1,\cdots,K\},$
$$\u_i \cdot \u_j =\begin{cases}
   \cos(\tfrac{(1-h)\pi}{2}) & \textrm{if } i\neq j, \\
   1 & \textrm{if } i = j.
\end{cases}$$
\end{theorem}

As shown in Equation~\eqref{eqn:generalbasis}, \newbasis is constructed as $\tau\P^k \x+(1-\tau)\u_k$ by integrating homophily basis and heterophily basis with a parameter $\tau\in[0,1]$. This integration yields enhanced adaptability of \newbasis to various graphs. Consequently,  \newbasis is better equipped to learn optimal weight parameters during model training compared with fixed polynomials. 

\spara{Estimation of homophily ratio} The accurate calculation of the homophily ratio $h$ relies on the label set of the entire graphs, which remains inaccessible. To circumvent this issue, we estimate $h$ through labels of training data, denoted as $\hat{h}$, as the input for \ours. We validate the estimation accuracy in the ablation study (Section~\ref{sec:ablation}) and robustness in Appendix~\ref{app:exp}. As shown, $\h$ can be efficiently estimated via a small proportion of labeled training nodes without compromising the performance of \ours. This fact signifies that the estimated homophily ratio $\hat{h}$ can serve as an ideal substitute for $h$. 

\spara{Time complexity} Algorithm~\ref{alg:basis} consists of $K$ iterations. In the $k$-th iteration, it takes $O(m+n)$ to calculate the orthonormal basis and $O(n)$ to update $\u_k$. Therefore, the total time complexity of Algorithm~\ref{alg:basis} is $O(K(m+n))$, \ie linear to propagation hops and input graph sizes.

\subsection{\newbasis for Graph Explanation}

Once \ours is well-trained, the derived \newbasis with the learned weights demystifies the spectral properties of graph signals. Specifically, given a graph $\G$ with node signal $\x$, \newbasis constitutes the spectrum $\{f(\x), f(\tau\P \x+(1-\tau)\u_1),\cdots, f(\tau\P^K\x+(1-\tau)\u_K)\}$. The $k$-th learned weight $\w_k$ from the weight vector $\w\in \R^{K+1}$ acts as the {\em amplification factor} of the signal in frequency $f(\tau\P^k\x+(1-\tau)\u_k)$. Consequently, the weight vector $\w$ discloses the significance of each frequency component in $\G$, thereby unfolding the distribution of graph signals and offering insights into the hidden spectral property. 

As validated in Section~\ref{sec:spectrum}, the experimental results (Figure~\ref{fig:datasetspectrum}) demonstrate the superior capability of \newbasis to capture the spectral characteristics of graphs across diverse heterophily degrees. Akin to the Fourier transform in signal processing, \newbasis manifests its potential as a promising approach for graph explanation.

\section{Theoretical Analysis}\label{sec:theory}

Every pair of vectors $(\u_i,\u_j)$ from the heterophily basis form the angle of $\theta$. Among them, $\u_0 \gets \tfrac{\x}{\|\x\|}$ is the normalized input node signal $\x$. In this regard, any heterophily vector $\u_i$ for $i\in \{1,2,\cdots,K\}$ can be obtained by rotating $\x$ with $\theta$ degree via a certain {\em rotation matrix} $\P_\theta$, \ie $\u_i:=\P_\theta\x$. Formally, the rotation matrix is defined as
\begin{definition}[Rotation matrix]\label{def:rotation}
Matrix $\P_\theta \in \R^{n\times n}$ is a rotation matrix if there exists a unitary matrix $\U\in \mathbb{C}^{n\times n}$ such that $\mathbf{R}(\theta):=\U^{-1}\P_\theta\U$ where $$\mathbf{R}(\theta)=\begin{pmatrix}\cos\theta &-\sin(\theta) & 0 & \cdots\\ \sin(\theta) & \cos\theta & 0 & \cdots \\ 0 & 0 & 1 & \cdots \\  &  & \cdots & \cdots \\ 0 & 0  & \cdots & 1 \end{pmatrix}.$$
\end{definition}
Therefore, {\em \ours} in Equation~\eqref{eqn:generalbasis} can be reformulated as 
\begin{equation}\label{eqn:convoform}
\z=\textstyle\sum^K_{k=0}\w_k \big((\tau\P^k+ (1-\tau)\P_{\theta,k})\x\big),   
\end{equation}
where $\P_{\theta,k}$ is the $k$-th rotation matrix and $\tau\P^k+ (1-\tau)\P_{\theta,k}$ works as the {\em convolutional matrix} of the $k$-th layer. 

\spara{Over-smoothing analysis} Consider a node feature matrix $\X\in \R^{n\times d}$ where $d$ is the feature dimension. The similarity of node representations at the $k$-th layer on graph $\G$ is measured by {\em Dirichlet energy} as $E(\G,\X^k)=\tfrac{1}{n}\sum_{v\in \V}\sum_{u\in \N_v}\|\X^k_v-\X^k_u\|^2_2$ where $\X^k$ is the resultant feature matrix at the $k$-th layer. Over-smoothing occurs if $E(\G,\X^k)\to 0$ for a sufficiently large $k$. In terms of this, we substantiate that \ours prevents over-smoothing. 

\begin{theorem}\label{thm:oversmoothing}
Given a node feature matrix $\X\in \R^{n\times d}$ on graph $\G$, let $\X^k=(\tau\P^k+ (1-\tau)\P_{\theta,k})\X$. It holds that $\lim_{K\to\infty}E(\G,\X^k) = (1-\tau)E(\G,\X)$.
\end{theorem}

\spara{Over-squashing analysis} Over-squashing~\cite{AlonY21, ToppingGC0B22} describes the phenomenon that exponentially growing information from distant nodes is squeezed into fixed-size vectors during message passing on graphs. Over-squashing is usually measured by the {\em Jacobian} of node representations~\cite{ToppingGC0B22, GiovanniGBLLB23}. Specifically, let $\z^{(k)}=\w_k \big((\tau\P^k+ (1-\tau)\P_{\theta,k})\x\big)$ be the node signal vector of the $k$-th step  in \ours and $k$ be the distance between nodes $u$ and $v$. Jacobian $|\partial \z^{(k)}_u / \partial \x_v|$ measures the sensitivity of information received at node $u$ to the signal of node $v$ when propagating $k$ distance. Over-squashing occurs if $|\partial \z^{(k)}_u / \partial \x_v| \le c \cdot (\P^k_M)_{uv}$ where $c$ is a constant and $\P_M$ is the message-passing matrix, \ie the sensitivity (node dependence) decays exponentially to propagation distance. However, we prove that \ours is able to mitigate over-squashing.

\begin{theorem}\label{thm:oversquashing}
Consider $\z^{(k)}$ as the $k$-th step node signal vector in \ours. Jacobian $|\partial \z^{(k)}_u / \partial \x_v|$ is independent of propagation step $k$.  
\end{theorem}

\section{Experiments}\label{sec:exp}

\spara{Datasets} We evaluate the performance of \ours on $6$ real-world datasets with varied homophily ratios. Specifically, the three citation networks~\citep{sen2008collective}, \ie Cora, Citeseer, and Pubmed, are homophily graphs with homophily ratios $0.81$, $0.73$, and $0.80$ respectively; the two Wikipedia graphs, \ie Chameleon and Squirrel and the Actor co-occurrence graph from WebKB3~\citep{PeiWCLY20} are heterophily graphs with homophily ratios $0.22$, $0.23$, and $0.22$ respectively. The dataset details are presented in Table~\ref{tbl:dataset} in Appendix~\ref{app:settings}.

\spara{Baselines} We compare \ours with $20$ baselines in two categories, \ie\ {\em polynomial filters} and {\em model-optimized methods}. For polynomial filters, we include monomial \SGC~\citep{WuSZFYW19}, \SIGN~\citep{frasca2020sign}, \revise{\ASGC~\cite{ChanpuriyaM22}}, \GPRGNN~\citep{ChienP0M21}, and \EvenNet~\citep{LeiWLDW22}, Chebyshev polynomial \ChebNet~\citep{DefferrardBV16} and its improved version \ChebNetII~\citep{HeWW22}, Bernstein polynomial \BernNet~\citep{he2021bernnet}, Jacobi polynomial \JacobiConv~\citep{WangZ22}, the orthogonal polynomial \OptBasisGNN~\citep{GuoW23} and learnable basis \Specformer~\citep{BoSWL23}. For model-optimized methods, they optimize the model architecture for improvement. We consider \GCN~\citep{KipfW17}, \GCNII~\citep{ChenWHDL20}, \GAT~\citep{VelickovicCCRLB18}, \MixHop~\citep{AbuElHaijaPKA19}, \HGCN~\citep{ZhuYZHAK20}, \LINKX~\citep{lim2021large}, \revise{\WRGAT~\citep{SureshBNLM21}}, \ACM~\citep{LuanHLZZZCP22}, and \GloGNN~\citep{LiZCSLLQ22}. 

\spara{Experiment Settings} There are two common data split settings, \ie $60\% / 20\% / 20\%$ and $48\% / 32\% / 20\%$ for training/validation/testing in the literature. Specifically, the polynomial filters are mostly tested in the previous setting~\citep{WangZ22,GuoW23,BoSWL23} while the model-optimized methods are normally evaluated in the latter\footnote{Please note that those model-optimized methods reuse the public data splits from~\citet{PeiWCLY20} which are actually in the splits of $48\%/32\%/20\%$ in the implementation.}~\citep{ZhuYZHAK20,LiZCSLLQ22,SongZWL23}.

\subsection{Node Classification Performance}\label{sec:classresults}\vspace{3mm}

\begin{table*}[!t]
\centering
 \caption{Accuracy (\%) compared with polynomial filters.}\label{tbl:polyfilter}\vspace{-1mm}
 \small
\begin{tabular}{@{}c|c|c|c|c|c|c@{}}
\toprule
\multicolumn{1}{c}{\bf Methods} & \multicolumn{1}{c}{\bf Cora} & \multicolumn{1}{c}{\bf Citeseer} & \multicolumn{1}{c} {\bf Pubmed} &\multicolumn{1}{c}{\bf Actor} & \multicolumn{1}{c} {\bf Chameleon}& \multicolumn{1}{c}{\bf Squirrel} \\ \midrule
\SGC             &	   86.83 $\pm$ 1.28 	&	   79.65 $\pm$ 1.02  	&	   87.14 $\pm$ 0.90    	 &	34.46 $\pm$ 0.67    &	   44.81 $\pm$ 1.20    &	 25.75 $\pm$ 1.07    \\
 \SIGN                   &	87.70 $\pm$ 0.69  	&	80.14 $\pm$ 0.87   	&	 89.09 $\pm$ 0.43      & 		41.22 $\pm$ 0.96      	&	60.92 $\pm$ 1.45         	&	45.59 $\pm$ 1.40    	\\
 \revise{\ASGC} & \revise{85.35 $\pm$ 0.98} & \revise{76.52 $\pm$ 0.36} & \revise{84.17 $\pm$ 0.24} &  \revise{33.41 $\pm$ 0.80} & \revise{71.38 $\pm$ 1.06} & \revise{57.91 $\pm$ 0.89} \\
 \GPRGNN               	&	   88.54 $\pm$ 0.67  	&	   80.13 $\pm$ 0.84  	&	   88.46 $\pm$ 0.31     &	   39.91 $\pm$ 0.62  	&	   67.49 $\pm$ 1.38     	&	   50.43 $\pm$ 1.89    \\
\EvenNet               &  87.77 $\pm$ 0.67   &	78.51 $\pm$ 0.63     	&\underline{90.87 $\pm$ 0.34} &  	40.36 $\pm$ 0.65       	&	67.02 $\pm$ 1.77         	&	52.71 $\pm$ 0.85  	\\
 \ChebNet              	&	   87.32 $\pm$ 0.92  	&	   79.33 $\pm$ 0.57  	&	   87.82 $\pm$ 0.24  	&	   37.42 $\pm$ 0.58   	&	   59.51 $\pm$ 1.25     	&	   40.81 $\pm$ 0.42     \\
 \ChebNetII             & 88.71  $\pm$ 0.93	    &	80.53  $\pm$ 0.79     	&	88.93  $\pm$ 0.29    &  	41.75 $\pm$ 1.07 &	71.37  $\pm$ 1.01        	&	57.72  $\pm$ 0.59    	\\
 \BernNet              	&	   88.51 $\pm$ 0.92  	&	   80.08 $\pm$ 0.75  	&	   88.51 $\pm$ 0.39    &	   41.71 $\pm$ 1.12 &	   68.53 $\pm$ 1.68     	&	   51.39 $\pm$ 0.92    	\\
 \JacobiConv           	&	   \underline{88.98 $\pm$ 0.72}  	&	   80.78 $\pm$ 0.79    	&	   89.62 $\pm$ 0.41  	&	   41.17 $\pm$ 0.64   	&	   74.20 $\pm$ 1.03     &	   57.38 $\pm$ 1.25   \\
 \OptBasisGNN           & 87.00 $\pm$ 1.55	    &	80.58  $\pm$ 0.82     	&	90.30  $\pm$ 0.19 &	\bf{42.39 $\pm$ 0.52} &	74.26 $\pm$ 0.74      	&	63.62  $\pm$ 0.76   	\\
 \Specformer    &	88.57 $\pm$ 1.01   & \underline{81.49 $\pm$ 0.94} & 87.73 $\pm$ 0.58    &  \underline{41.93 $\pm$ 1.04} & \underline{74.72 $\pm$ 1.29}    &  \underline{64.64 $\pm$ 0.81}	\\
 \ours      &	\revise{\bf{89.49 $\pm$ 1.35}}   &  \revise{\bf{81.39 $\pm$ 1.32}} & \revise{\bf{91.44 $\pm$ 0.50}}  & \revise{40.84 $\pm$ 1.21} &  \revise{\bf{75.75 $\pm$ 1.65}}  & \revise{\bf{67.40 $\pm$ 1.25}}    	\\ \bottomrule 
\end{tabular}
\end{table*}
\begin{table*}[!t]
\centering
 \caption{Accuracy (\%) compared with model-optimized methods.}\label{tbl:modelopt}\vspace{-1mm}
 \small
\begin{tabular}{@{}c|c|c|c|c|c|c@{}}
\toprule
\multicolumn{1}{c}{\bf Methods} & \multicolumn{1}{c}{\bf Cora} & \multicolumn{1}{c}{\bf Citeseer} & \multicolumn{1}{c} {\bf Pubmed} &\multicolumn{1}{c}{\bf Actor} & \multicolumn{1}{c} {\bf Chameleon}& \multicolumn{1}{c}{\bf Squirrel} \\ \midrule
 \GCN   &	   86.98 $\pm$ 1.27  	&	   76.50 $\pm$ 1.36  	&	   88.42 $\pm$ 0.50    	&	   27.32 $\pm$ 1.10    	&	   64.82 $\pm$ 2.24     	&	   53.43 $\pm$ 2.01    	\\
\GCNII  &	\underline{88.37 $\pm$ 1.25}   & \underline{77.33 $\pm$ 1.48} & \underline{90.15 $\pm$ 0.43} & 37.44 $\pm$ 1.30 & 63.86 $\pm$ 3.04 & 38.47 $\pm$ 1.58   \\
\GAT    &	87.30 $\pm$ 1.10   &  76.55 $\pm$ 1.23 & 86.33 $\pm$ 0.48 & 27.44 $\pm$ 0.89 & 60.26 $\pm$ 2.50  & 40.72 $\pm$ 1.55  \\
\MixHop  &	87.61 $\pm$ 0.85   & 76.26 $\pm$ 1.33 & 85.31 $\pm$ 0.61 & 32.22 $\pm$ 2.34 &  60.50 $\pm$ 2.53 & 43.80 $\pm$ 1.48   \\
\HGCN   &	87.87 $\pm$ 1.20   & 77.11 $\pm$ 1.57 & 89.49 $\pm$ 0.38 & 35.70 $\pm$ 1.00 & 60.11 $\pm$ 2.15 & 36.48 $\pm$ 1.86   \\
\LINKX  &	84.64 $\pm$ 1.13   & 73.19 $\pm$ 0.99 & 87.86 $\pm$ 0.77 & 36.10 $\pm$ 1.55 & 68.42 $\pm$ 1.38 & 61.81 $\pm$ 1.80   \\
\revise{\WRGAT} & \revise{88.20 $\pm$ 2.26} & \revise{76.81 $\pm$ 1.89} & \revise{88.52 $\pm$ 0.92} & \revise{36.53 $\pm$ 0.77} & \revise{65.24 $\pm$ 0.87} & \revise{48.85 $\pm$ 0.78} \\
\ACM    &	87.91 $\pm$ 0.95   & 77.32 $\pm$ 1.70 & 90.00 $\pm$ 0.52 & 36.28 $\pm$ 1.09 & 66.93 $\pm$ 1.85  & 54.40 $\pm$ 1.88   \\
\GloGNN &	88.33 $\pm$ 1.09   & 77.22 $\pm$ 1.78 & 89.24 $\pm$ 0.39 & \underline{37.70 $\pm$ 1.40} & \underline{71.21 $\pm$ 1.84}  & \underline{57.88 $\pm$ 1.76}   \\
\ours   &	\revise{\bf{89.12 $\pm$ 0.87}}  & \revise{\bf{80.28 $\pm$ 1.31}} & \revise{\bf{90.19 $\pm$ 0.41}} & \revise{\bf{37.79 $\pm$ 1.11}} & \revise{\bf{73.66 $\pm$ 2.44}} & \revise{\bf{64.26 $\pm$ 1.46}}   \\ \bottomrule 
\end{tabular}

\vspace{-2mm}
\end{table*} 

Table~\ref{tbl:polyfilter} and Table~\ref{tbl:modelopt} report the accuracy scores associated with standard deviations for \ours, tested polynomial filters, and model-optimized methods for node classification. For ease of exposition, we highlight the {\em highest} accuracy score in bold and underline the {\em second highest} score for each dataset. 

As shown, our method \ours consistently achieves the highest accuracy scores on both the homophily datasets and heterophily datasets, except in one case on Actor in Table~\ref{tbl:polyfilter}. \ours exhibits explicit performance advantages over both SOTA polynomial filter \Specformer and SOTA model-optimized method \GloGNN for the majority of cases. In particular, the performance improvements are remarkably significant on the two heterophily datasets Chameleon and Squirrel. Specifically, the corresponding performance gains reach up to \revise{$1.03\%$} and \revise{$2.76\%$} as shown in Table~\ref{tbl:polyfilter} and \revise{$2.45\%$} and \revise{$6.38\%$} as shown in Table~\ref{tbl:modelopt} respectively. It is worth mentioning that the computation time of \newbasis is linear to graph sizes and propagation hops. The superior performance of \ours significantly confirms the outstanding effectiveness and universality of \newbasis.

\subsection{Signal Spectrum for Graph Explanation}\label{sec:spectrum}

\begin{figure*}
\centering
\begin{small}
\subfloat[{Cora}]{
\begin{tikzpicture}
\begin{axis}[
    ybar=2pt,
    height=\columnwidth/2.5,
    width=\columnwidth/1.7,            
    bar width=0.05cm,
    enlarge x limits=true,
    ymin=0, ymax=0.35,
    xmin=0.38, xmax=0.91,         
    xlabel={{\em Frequency f(x)}},
    ylabel={\em Weights $\w$},
    xlabel style={yshift=0.1cm},
    ylabel style={yshift=-0.1cm},    
    xtick={0.4,0.5,0.6,0.7,0.8,0.9},
    xticklabel style={font=\scriptsize},
    ]
    \addplot[fill=cyan] coordinates {(0.9049213, 0.28665602) (0.7512426, 0.21770808)(0.5989015, 0.31996652)(0.55581844, 0.22022095)(0.46699378, 0.25615853)(0.46476635, 0.17635748)(0.4138301, 0.17979105)(0.42232922, 0.12524088)(0.39268094, 0.13313477)(0.40258533, 0.09199194)(0.38496265, 0.09186243)};    
\end{axis}
\end{tikzpicture}
\label{subfig:cora}
}\hspace{2mm}
\subfloat[{Citeseer}]{
\begin{tikzpicture}
\begin{axis}[
    ybar=2pt,
    height=\columnwidth/2.5,
    width=\columnwidth/1.7,            
    bar width=0.05cm,
    enlarge x limits=true,
    ymin=0.01, ymax=0.1,
    xmin=0.39, xmax=0.85,
    xlabel={{\em Frequency f(x)}},
    ylabel={\em Weights $\w$},
    yticklabel={\pgfmathprintnumber[fixed,precision=2]{\tick}},
    xtick={0.4,0.5,0.6,0.7,0.8},
    xlabel style={yshift=0.1cm},
    ylabel style={yshift=-0.1cm},
    xticklabel style={font=\scriptsize},
    ]
    \addplot[fill=cyan] coordinates {(0.8306937, 0.015354116) (0.39249885, 0.08794039)(0.7361769, 0.06907876)(0.4353334, 0.074572735)(0.7140484, 0.07314714)(0.46884233, 0.08724001)(0.7073295, 0.075021446)(0.49079067, 0.07216103)(0.7056873, 0.0796282)(0.50604355, 0.079077415)(0.705987, 0.056096457)};    
\end{axis}
\end{tikzpicture}
\label{subfig:citeseer}
}\hspace{2mm}
\subfloat[{Pubmed}]{
\begin{tikzpicture}
\begin{axis}[
    ybar=2pt,
    height=\columnwidth/2.5,
    width=\columnwidth/1.7,            
    bar width=0.05cm,
    enlarge x limits=true,
    ymin=0.01, ymax=0.350,
    xmin=0.3, xmax=0.8,         
    xlabel={{\em Frequency f(x)}},
    ylabel={\em Weights $\w$},
    xtick={0.3, 0.4,0.5,0.6,0.7,0.8},
    xticklabel style={font=\scriptsize},
    xlabel style={yshift=0.1cm},
    ylabel style={yshift=-0.1cm},
    ]
    \addplot[fill=cyan] coordinates {(0.79359525, 0.3262818)(0.60967284, 0.007923197)(0.45221677, 0.17476733)(0.5022794, 0.03927678)(0.37954324, 0.1305862)(0.46210948, 0.03577217)(0.3504402, 0.09895706)(0.43751678, 0.029734226)(0.3378904, 0.076884076)(0.4203169, 0.02118163)(0.3325249, 0.058505483)};    
\end{axis}
\end{tikzpicture}
\label{subfig:pubmed}
}
\vspace{-1mm}
\subfloat[{Actor}]{
\begin{tikzpicture}
\begin{axis}[
    ybar=2pt,
    height=\columnwidth/2.5,
    width=\columnwidth/1.7,            
    bar width=0.05cm,
    enlarge x limits=true,
    ymin=-0.04, ymax=0.35,
    xmin=0.65, xmax=1,         
    xlabel={{\em Frequency f(x)}},
    ylabel={\em Weights $\w$},
    xtick={0.7,0.8,0.9,1.0},
    xticklabel style={font=\scriptsize},
    xlabel style={yshift=0.1cm},
    ylabel style={yshift=-0.1cm},
    ]
    \addplot[fill=cyan] coordinates {(0.99351937, 0.3371407)(0.7944303, 0.25218594)(0.743898, 0.08607966)(0.7694793, 0.005307203)(0.6680614, -0.0017618219)(0.69484097, -0.008277266)(0.7032763, -0.014003986)(0.6826355, 0.0028506124)(0.72351223, 0.016178958)(0.7077483, 0.013457422)(0.7282677, 0.02094628)}; 
    \draw[black, dashed] ( 0.65, 0) -- ( 1, 0);    
\end{axis}
\end{tikzpicture}
\label{subfig:actor}}\hspace{2mm}
\subfloat[{Chameleon}]{
\begin{tikzpicture}
\begin{axis}[
    ybar=2pt,
    height=\columnwidth/2.5,
    width=\columnwidth/1.7,            
    bar width=0.05cm,
    enlarge x limits=true,
    ymin=-0.15, ymax=0.2,
    xmin=0.78, xmax=1.0,         
    xlabel={{\em Frequency f(x)}},
    ylabel={\em Weights $\w$},
    xtick={0.8,0.85, 0.9, 0.95, 1.0},
    xticklabel style={font=\scriptsize},
    xlabel style={yshift=0.1cm},
    ylabel style={yshift=-0.1cm},
    ]
    \addplot[fill=cyan] coordinates {(0.99189526, -0.010718918)(0.8896081, -0.032063093)(0.8741272, 0.17320552)(0.8245465, 0.0626653)(0.8099159, -0.1427136)(0.78005356, 0.03798501)(0.78844684, 0.09077762)(0.7926478, -0.04823897)(0.7953111, -0.08292176)(0.78577465, -0.016427025)(0.7937439, 0.095499545)};
    \draw[black, dashed] ( 0.75, 0) -- ( 1, 0);    
\end{axis}
\end{tikzpicture}
\label{subfig:chameleon}
}\hspace{2mm}
\subfloat[{Squirrel}]{
\begin{tikzpicture}
\begin{axis}[
    ybar=2pt,
    height=\columnwidth/2.5,
    width=\columnwidth/1.7,            
    bar width=0.05cm,
    enlarge x limits=true,
    ymin=-1.5, ymax=1.4,
    xmin=0.35, xmax=1.0,         
    xlabel={{\em Frequency f(x)}},
    ylabel={\em Weights $\w$},
    xtick={0.3, 0.4, 0.5, 0.6, 0.7, 0.8, 0.9, 1.0},
    xticklabel style={font=\scriptsize},
    xlabel style={yshift=0.1cm},
    ylabel style={yshift=-0.1cm},
    ]
    \addplot[fill=cyan] coordinates {(0.9874364, 0.14482574)(0.75930494, 0.08507316)(0.44606453, 1.217764)(0.58042026, -0.13832489)(0.35057908, -1.4312787)(0.47658646, 0.38738832)(0.36176968, 0.68456066)(0.42023587, -0.5975129)(0.36092305, -1.290962)(0.35031483, 0.7090253)(0.3487408, 0.6671545)};
    \draw[black, dashed] ( 0.3, 0) -- ( 1, 0);    
\end{axis}
\end{tikzpicture}
\label{subfig:squirrel}
}
\end{small}\vspace{-1mm}
\captionof{figure}{Spectrum distribution of trained \newbasis.} \label{fig:datasetspectrum}
\end{figure*}
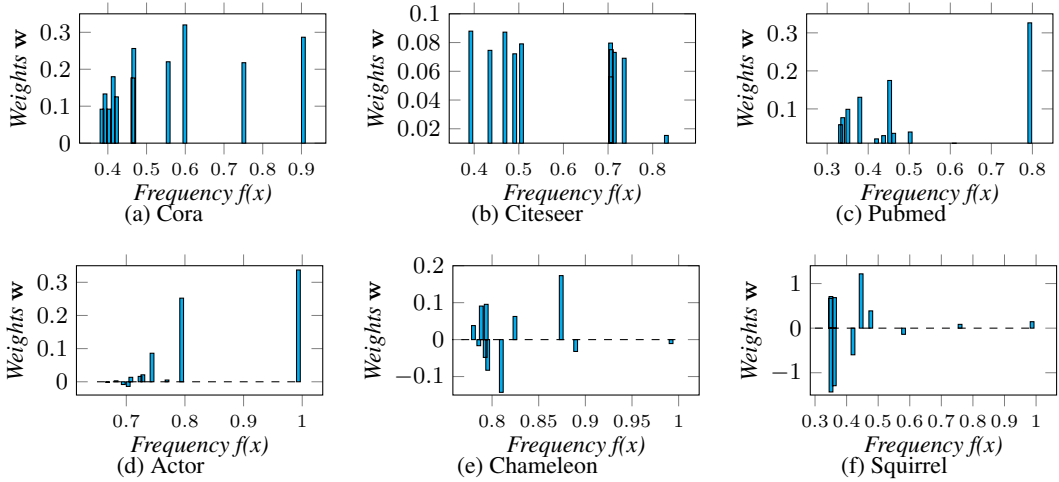

The superior performance of \ours explicitly implies the outstanding capability of \newbasis to capture the spectral characteristics of graphs. For a better demonstration, we first calculate the {\em spectral signal frequency} of each basis vector for all $d$-dimensions, resulting in $d$ spectrum of length $K+1$. We then average the spectrum and associate it with the learned weights $\w$ accordingly, where weights $\w \in \R^{K+1}$ of \newbasis are trained for each dataset. The spectrum distributions of the trained \newbasis for the $6$ datasets are plotted in Figure~\ref{fig:datasetspectrum}.

Recall that signals in specific frequencies with weights in large absolute are enhanced while signals with small weights are suppressed. As displayed, the majority of signals of the three homophily datasets lie within the relatively low-frequency intervals as expected, \eg $[0.3,0.5]$. We also observe some minor high-frequency information which also provides insightful information for node classification~\citep{KlicperaWG19, Chen0XYKRYF19, balcilar2020bridging} and node distinguishability~\cite{LuanHXLZCFLP23}. On the contrary, \newbasis of the three heterophily datasets aims to remove low-frequency signals with negative weights and preserve high-frequency information. The distinct spectrum distributions of \newbasis disclose the unique spectral characteristics of each dataset. These results highlight the capability of \newbasis as a new method to analyze graphs with varying heterophily degrees in the spectral domain with enriched understanding. 

\subsection{Ablation Studies}\label{sec:ablation}

\begin{figure*}[!t]
  \centering
  \begin{minipage}[t]{0.48\textwidth}
    \centering
    \begin{tikzpicture}[scale=1,every mark/.append style={mark size=1.5pt}]
        \begin{axis}[
            height=\columnwidth/1.8,
            width=\columnwidth/1,
            ylabel={\em Accuracy gap} (\%),
            xlabel={\em Varying homo. ratio $h$},
            xmin=0.5, xmax=8.5,
            ymin=-0.2, ymax=4,
            xtick={1,2,3,4,5,6,7,8},
            xticklabel style={font=\scriptsize},
            yticklabel style={font=\footnotesize},
            xticklabels={0.13,0.2, 0.3, 0.4, 0.5, 0.6, 0.7, 0.81},
            ylabel style={yshift=-0.1cm},
            legend columns=3,
            legend style={fill=none,font=\footnotesize,at={(0.35,1.1)},anchor=center,draw=none},
        ]
        \addplot[line width=0.25mm,mark=triangle,color=cyan]
        coordinates {
            (1, 0) (2, 0) (3, 1.38) (4, 6.86) (5, 10.59) (6,17.76) (7,29.78) (8,35.76) }; 
        \addplot[line width=0.25mm,mark=diamond,color=blue]
        coordinates {
            (1, 1.6) (2, 0.43) (3,0) (4, 0) (5, 0.52) (6, 0.51) (7, 0.13) (8,0)}; 
        \addplot[line width=0.25mm,mark=o,color=red]
        coordinates {
            (1, 0.73) (2, 1.55) (3,2.69) (4, 1.79) (5, 1.48) (6, 1.4) (7, 0.15) (8, 0.71) }; 
        \legend{\HetFilter, \HomFilter, \OrtFilter}
        \end{axis}
    \end{tikzpicture}
    \caption{Accuracy gaps of the three variants from \ours on $\G_s$ across varying $h$.} \label{fig:varyingh}
  \end{minipage}
  \begin{minipage}[t]{0.48\textwidth}
    \centering
   \begin{tikzpicture}[scale=1,every mark/.append style={mark size=1.5pt}]
    \begin{axis}[
        height=\columnwidth/1.8,
        width=\columnwidth/1,
        ylabel={\em Accuracy (\%)},
        xlabel={\em Varying $\tau$},
        xmin=0.5, xmax=10.5,
        ymin=55, ymax=92,
        xtick={1,2,3,4,5,6,7,8,9,10},
        xticklabel style = {font=\footnotesize},
        yticklabel style = {font=\footnotesize},
        xticklabels={ 0.1,0.2,0.3,0.4,0.5,0.6,0.7,0.8,0.9,1},
        ylabel style={yshift=-0.1cm},
        legend columns=2,
        legend style={fill=none,font=\footnotesize,at={(0.35,1.1)},anchor=center,draw=none},
    ] 
    \addplot [line width=0.25mm, color=blue, error bars/.cd, y dir=both, y explicit] 
    table [x=x, y=y, y error=y-err]{%
    x   y   y-err
    1	70.53	2.31
    2	72.02	1.34
    3	74.24	2.19
    4	77.54	0.88
    5	78.90	1.61
    6	81.87	1.36
    7	84.56	2.11
    8	86.93	1.04
    9	89.39	1.36
    10  89.11   1.48
    };
    \addplot [line width=0.25mm, color=red, error bars/.cd, y dir=both, y explicit] 
    table [x=x, y=y, y error=y-err]{%
    x   y   y-err
    1	66.75	1.05
    2	66.87	1.70
    3	65.85	1.51
    4	66.62	0.85
    5	66.66	0.96
    6	67.09	1.08
    7	67.01	1.25
    8	66.07	0.65
    9	65.40	1.39
    10  58.65   1.15
    };
    \legend{\textsf{Cora}, \textsf{Squirrel}}
    \end{axis}
\end{tikzpicture}
\caption{Accuracy (\%) with varying $\tau$.} \label{fig:varyingtau}
  \end{minipage}\vspace{-2mm}
\end{figure*}

\spara{Universality of \newbasis} To verify the effectiveness of \newbasis, we devise $3$ variants of \ours with different polynomial bases. To this end, we alter \ours by changing \newbasis into 1) a filter simply using the heterophily basis (setting $\tau=0$) denoted as \HetFilter, 2) a filter simply using the homophily basis (setting $\tau=1$) denoted as \HomFilter, and 3) a filter using the orthonormal basis ($\{\v_0, \v_1, \cdots, \v_k\}$) denoted as \OrtFilter. For easy control, we generate a synthetic dataset $\G_s$ by adopting the graph structure and label set of Cora. Without loss of generality, we generate a random one-hot feature vector in $100$ dimensions for each node. To vary the homophily ratios of $\G_s$, we permute nodes in a random sequence and randomly reassign node labels progressively, resulting in homophily ratios in $\{0.13\footnote{Note that this is the smallest homophily ratio we can possibly acquire by random reassignments.}, 0.20, 0.30, 0.40, 0.50, 0.60, 0.70, 0.81\}$. 

The performance advantage gaps of \ours over the three variants are presented in Figure~\ref{fig:varyingh}. We omit the results of \HetFilter beyond $h\ge 0.3$ since the corresponding performance gaps become significantly larger, which is as expected since the heterophily basis is incapable of tackling homophily graphs. In particular, the performance advantage of \ours over \HomFilter gradually decreases as $h$ grows larger. In contrast, the performance gaps of \OrtFilter from \ours peak at $h=0.3$ with a notable shortfall and then erratically decrease, ending with an accuracy gap of $0.71\%$ at $h=0.81$. The fluctuation of \OrtFilter states the inferiority of the orthonormal basis over \newbasis.

\spara{Sensitivity of $\tau$} To explore the sensitivity of \ours to the hyperparameter $\tau$, we vary $\tau$ in $\{0,0.1,\cdots,0.9,1\}$ and test \ours on the strong homophily dataset Cora and the strong heterophily dataset Squirrel. Figure~\ref{fig:varyingtau} plots the performance development with varying $\tau$. As displayed, \ours prefers the homophily basis on Cora, and the performance peaks at $\tau=0.9$. On the contrary, the performance of \ours slightly fluctuates when $\tau \le 0.7$ and then moderately decreases along the increase of $\tau$ on Squirrel. When $\tau=1.0$, the accuracy score drops sharply since only the homophily basis is utilized in this scenario.

\begin{table*}[!t]
\centering
 \caption{Estimated homophily ratios $\hat{h}$.} \label{tbl:EstHomRatio}\vspace{-1mm} 
    \setlength{\tabcolsep}{0.5em}
 \small
 \renewcommand\arraystretch{1.3}
\begin{tabular} {@{}c|cccccc@{}}
\toprule
{\bf Dataset}  & \multicolumn{1}{c}{Cora} & \multicolumn{1}{c}{Citeseer} & \multicolumn{1}{c} {Pubmed} &\multicolumn{1}{c}{Actor}& \multicolumn{1}{c} {Chameleon}& \multicolumn{1}{c}{Squirrel}\\ \midrule 
{True $h$}  & 0.81  & 0.74  & 0.80  & 0.22  & 0.23  & 0.22  \\ \hline
{$\hat{h}_1$}  & 0.82 $\pm$ 0.01 & 0.70 $\pm$ 0.01 & 0.79 $\pm$ 0.01 & 0.21 $\pm$ 0.004 & 0.24 $\pm$ 0.01 & 0.22 $\pm$ 0.01 \\ \hline
{$\hat{h}_2$} & 0.82 $\pm$ 0.01 & 0.69 $\pm$ 0.01 & 0.79 $\pm$ 0.01 & 0.21 $\pm$ 0.004 & 0.24 $\pm$ 0.01 & 0.22 $\pm$ 0.01 \\ \bottomrule
\end{tabular}
\end{table*}

\spara{Homophily Ratio Estimation} Notice that the homophily ratio is estimated on the training set from each random split. To demonstrate the accuracy of the estimation, we present the estimated homophily ratio $\hat{h}$ with standard deviation averaged over the $10$ estimations for all datasets in Table~\ref{tbl:EstHomRatio}. In particular, $\hat{h}_1$ and $\hat{h}_2$ represent the averaged homophily ratio estimations for data splits settings of $60\%/20\%/20\%$ for polynomial filters and $48\%/32\%/20\%$ for model-optimized methods respectively. As shown in Table~\ref{tbl:EstHomRatio}, the estimated values $\hat{h}_1$ and $\hat{h}_2$ closely align with the actual homophily ratio $h$. There is a difference within a $2\%$ range across all datasets, except for Citeseer. As validated from the superior performance of \ours in Table~\ref{tbl:polyfilter} and Table~\ref{tbl:modelopt}, $\hat{h}$ is an ideal substitute for $h$ as the input of \ours without compromising the performance. 

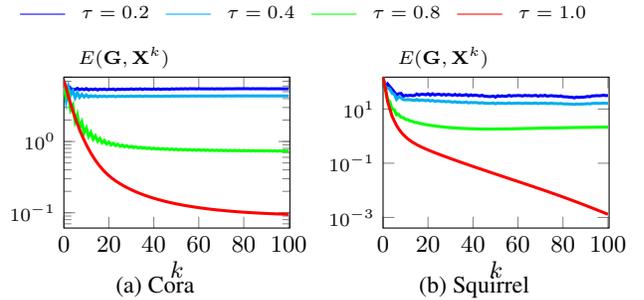
\begin{figure}[!bpt]
\centering
\begin{small}
\hspace{-1mm}
\begin{tikzpicture}
\begin{customlegend}[legend columns=4,legend style={align=center,draw=none,column sep=1ex, font=\scriptsize},
        legend entries={\bf $\tau=0.2$, $\tau=0.4$, $\tau=0.8$, $\tau=1.0$}]
        \addlegendimage{color=blue,solid,line legend}
        \addlegendimage{color=cyan,solid,line legend}
        \addlegendimage{color=green,solid,line legend}
        \addlegendimage{color=red,solid,line legend}        
        \end{customlegend}
\end{tikzpicture}\vspace{-3mm}
\subfloat[Cora]{
\begin{tikzpicture}[scale=1,every mark/.append style={mark size=1.5pt}]
    \begin{axis}[
        height=\columnwidth/2.3,
        width=\columnwidth/1.8,
        ylabel={\em $E(\G,\X^k)$},
        xlabel={\em $k$},
        xlabel style={yshift=0.1cm},
        xmin=0, xmax=100.5,
        ymin=0, ymax=8,
        xtick={0,20,40,60,80,100},
        yticklabel style = {font=\scriptsize},
        ylabel style={font=\footnotesize},
        ymode=log,
        log basis y={10},
        every axis y label/.style={font=\scriptsize,at={(current axis.north west)},right=8mm,above=0mm},
    ] 
    \addplot[line width=0.4mm,color=blue] file[skip first]{cora2.txt}; 
    \addplot[line width=0.4mm,color=cyan] file[skip first]{cora4.txt}; 
    \addplot[line width=0.4mm,color=green] file[skip first]{cora8.txt}; 
    \addplot[line width=0.4mm,color=red] file[skip first]{cora10.txt}; 
    \end{axis}
\end{tikzpicture}}
\subfloat[Squirrel]{
\begin{tikzpicture}[scale=1,every mark/.append style={mark size=1.5pt}]
    \begin{axis}[
        height=\columnwidth/2.3,
        width=\columnwidth/1.8,
        ylabel={\em $E(\G,\X^k)$},
        xlabel={\em $k$},
        xlabel style={yshift=0.1cm},
        xmin=0, xmax=100.5,
        ymin=0, ymax=150,
        xtick={0,20,40,60,80,100},
        yticklabel style = {font=\scriptsize},
        ylabel style={font=\footnotesize},
        ymode=log,
        log basis y={10},
        every axis y label/.style={font=\scriptsize,at={(current axis.north west)},right=8mm,above=0mm},
    ] 
    \addplot[line width=0.4mm,color=blue] file[skip first]{squirrel2.txt}; 
    \addplot[line width=0.4mm,color=cyan] file[skip first]{squirrel4.txt}; 
    \addplot[line width=0.4mm,color=green] file[skip first]{squirrel8.txt}; 
    \addplot[line width=0.4mm,color=red] file[skip first]{squirrel10.txt}; 
    \end{axis}
\end{tikzpicture}\hspace{0mm}}
\end{small}
\caption{Dirichlet energy $E(\G,\X^k)$ with varying $k$.} \label{fig:dirichlet}\vspace{-4mm}
\end{figure}

\spara{Prevention of Over-smoothing} We calculate the Dirichlet energy $E(\G,\X^k)$ of node representations $\X^k$ on dataset Cora and Squirrel by varying $k\in [1,100]$, demonstrating the capability of \ours to prevent the over-smoothing issue. Specifically, we set $\tau\in \{0.2,0.4,0.8,1.0\}$ and plot the corresponding $E(\G,\X^k)$ in Figure~\ref{fig:dirichlet}. Note that \ours utilizes solely homophily basis when $\tau=1$. Observe that the Dirichlet energy $E(\G,\X^k)$ approaches $0$ on both datasets when $\tau=1$ for sufficiently large $k$. This is attributed to the over-smoothing issue within the homophily basis, as elucidated in Theorem~\ref{thm:homo}. In contrast, when $\tau>0$, $E(\G,\X^k)$ remains a constant value larger than $0$ as $k$ increases. Furthermore, $E(\G,\X^k)$ increases with a decrease in $\tau$, aligning with the findings in Theorem~\ref{thm:oversmoothing}.

\section{Related Work}\label{sec:relatedwork}

\spara{Polynomial filters} As the seminal work, \citet{bruna2014spectral} proposes to generalize convolutional neural networks to graphs based on the spectrum of the graph Laplacian. Subsequently, \ChebNet~\citep{DefferrardBV16} utilizes a $K$-order truncated Chebyshev polynomial and provides a $K$-hop localized filtering capability. \GPRGNN~\citep{ChienP0M21} simply adopts monomials instead and applies the generalized PageRank~\citep{0005CM19} scores as the coefficients to measure node proximity. Meanwhile, \citet{ZhengZG0L0M21} exploits framelet transform to enhance the performance. To enhance controllability and interpretability, \BernNet~\citep{he2021bernnet} employs nonnegative Bernstein polynomials as the basis. Later, \citet{WangZ22} examines the expressive power of existing polynomials and proposes \JacobiConv by leveraging Jacobi polynomials~\citep{askey1974positive}, achieving better adaptability to underlying graphs. Subsequently, \citet{HeWW22} revisits \ChebNet and pinpoint the over-fitting issue in Chebyshev approximation. To address the issue, they turn to Chebyshev interpolation and propose \ChebNetII. Recently, polynomial filter~\OptBasisGNN~\citep{GuoW23} orthogonalizes the polynomial basis to maximize convergence speed. Instead of using fixed-order polynomials, \Specformer~\citep{BoSWL23} resorts to Transformer~\citep{VaswaniSPUJGKP17} to derive learnable bases for each feature dimension. However, it requires conducting eigendecomposition with the cost of $O(n^3)$, rendering it impractical for large social graphs. Nonetheless, the above polynomial filters do not take the varying heterophily degrees of graphs into consideration when utilizing polynomials, which leads to suboptimal empirical performance, as verified in our experiments.

\spara{Model-optimized GNNs} One common technique in model design is to combine both low-pass and high-pass filters. \textsf{GNN-LF/HF}~\citep{ZhuWSJ021} devises variants of the Laplacian matrix to construct a low-pass and high-pass filter respectively. \ACM~\citep{LuanHLZZZCP22} trains both low-pass and high-pass filters in each layer and then integrates their embeddings adaptively. Another strategy of model design is to extract homophily from both local and global graph structures. Specifically, \HGCN~\citep{ZhuYZHAK20} utilizes embeddings and intermediate representations of nodes at various distances, including central nodes, neighbors, and distant nodes. Similarly, \GloGNN~\citep{LiZCSLLQ22} trains a coefficient matrix in each layer to measure the correlations between nodes to aggregate homophilous nodes globally. To explicitly capture the relations between distant nodes, \WRGAT~\citep{SureshBNLM21} leverages the graph rewiring~\citep{ToppingGC0B22, KarhadkarBM23} technique by constructing new edges with weights to measure node proximity. Additionally, there are GNNs handling heterophily graphs from other aspects. \LINKX~\citep{lim2021large} learns embeddings from both node features and graph structure simultaneously. \OrderedGNN~\citep{SongZWL23} constructs the hierarchy structure of neighbors and then constrains the neighbor nodes from specific hops into the same blocks, without mixing features across hops.

\section{Conclusion}\label{sec:con}

In this paper, we propose a universal polynomial basis \newbasis by incorporating the graph heterophily degrees in the premise of thorough theoretical analysis. Utilizing \newbasis, we devise a general graph filter \ours. Meanwhile, \ours is capable of effectively preventing over-smoothing and mitigating over-squashing by optimizing the convolution matrix. Comprehensive evaluation supports the superiority of \ours on a diverse range of both real-world and synthetic datasets, which validates the effectiveness of \newbasis for graphs with varying heterophily degrees. The \newbasis also provides as a graph explainer for the spectral distribution of graph signals.

\section*{Impact Statement}

This paper presents work whose goal is to advance the field of 
Machine Learning. There are many potential societal consequences of our work, none which we feel must be specifically highlighted here.

\bibliography{ref}
\bibliographystyle{icml2024}

\newpage
\appendix
\onecolumn

\section{Notations and Tables}\label{app:notations}

\begin{table*}[!ht]
    \centering
    \caption{Frequently used notations.} \label{tbl:notations}
    \setlength{\tabcolsep}{0.3em} 
    \renewcommand{\arraystretch}{1.2}
    \renewcommand{\aboverulesep}{0pt}
    \renewcommand{\belowrulesep}{0pt}
\begin{tabular}{@{}cl@{}}\toprule
    \textbf{Notation} & \multicolumn{1}{c}{\textbf{Description}} \\ \midrule
    $\G=(\V,\E)$ & an undirected and connected graph with node set $\V$ and edge set $\E$\\ 
    $n,m$ & the numbers of nodes and edges in $\G$ respectively \\ 
    $[n]$ & the integer set $\{1,2,\cdots, n\}$ \\    
    $\N_u, d_u$ & the neighbor set of node $u$ and its degree\\ 
    $\X \in \R^{n\times d}$ & the feature matrix / node signals in $d$-dimension \\
    $\Y\in \mathbb{N}^{n\times |\C|}$ & the one-hot label matrix with the set of node labels $\C$ \\
    $\A, \TA$ & the adjacency matrix of $\G$ without and with self-loops respectively\\
    $\D, \TD$ & the diagonal degree matrix of $\G$ without and with self-loops respectively\\
    $\L, \hat{\L}$ & the normalized Laplacian matrix of $\G$ without and with self-loops respectively\\      
    $\P, \P_{\theta,k}$ & the propagation matrix and the $k$-th rotation matrix\\
    $\lambda_i$ & the $i$-th eigenvalue of Laplacian matrix $\L$ \\
    $K$ & the number of propagation hop \\
    $\w\in \R^{K+1}$ & the learnable of weight vector of polynomial basis \\
    $\U, \bLambda$ & the eigenvector matrix and diagonal eigenvalue matrix of $\L$ \\    
    $\Z\in \R^{n\times d}$ & the filtered node representations / signals in $d$-dimension  \\ 
    $h, \hat{h}$ & the homophily ratio of graphs defined in Definition~\ref{def:homo} and its estimation \\
    $f(\cdot)$ & the spectral signal frequency function defined in Definition~\ref{def:frequency} \\
    \bottomrule
\end{tabular}
\end{table*} 

\begin{table*}[h]
\caption{Polynomial Graph Filters} \label{tbl:pgf}\vspace{-2mm}
\small
\vspace{-1mm}
\renewcommand{\arraystretch}{1.2}
\centering
\begin{tabular}{@{}l|l|l|l}
\toprule
{} & Poly. Basis & Graph Filter $\g_\w(\lambda)$ & Prop. Matrix $\P$  \\ \midrule
\ChebNet~\citep{DefferrardBV16} & Chebyshev & $\sum^K_{k=0}\w_k T_k(\hat{\lambda})$ & $2\L/\lambda_{max}-\I$ \\ 
\GPRGNN~\citep{ChienP0M21}  & Monomial & $\sum^K_{k=0}\w_k(1-\Tlambda)^k$ & $\I-\hat{\L}$ \\
\BernNet~\citep{he2021bernnet} & Bernstein & $\sum^K_{k=0}\frac{\w_k}{2^K}\binom{K}{k}(2-\lambda)^{K-k}\lambda^k$ & $\I-\frac{\L}{2}$ \\ 
\JacobiConv~\citep{WangZ22} & Jacobi & $\sum^K_{k=0}\w_k\P^{a,b}_k(1-\lambda)$ & $\I-\L$ \\ 
\OptBasisGNN~\citep{GuoW23} & Orthonormal & --- & $\I-\L$ \\  \bottomrule
\end{tabular}
\vspace{-1mm}
\end{table*}

\section{Proofs}\label{app:proofs}

\subsection{Proof of Proposition~\ref{pro:frequencybound}}
\begin{proof}
Given a graph $\G=(\V,\E)$ and a normalized signal $\x \in \R^n$ on $\G$, the signal frequency $f(x)$ of $\x$ is {\large$$f(\x)=\tfrac{\x^\top\L\x}{2}=\tfrac{\sum_{{\langle u,v\rangle}\in \E}(\x_u-\x_v)^2}{2\sum_{u\in \V}\x^2_u d_u}$$} where $\L$ is the normalized Laplacian matrix defined on $\G$. Therefore, $f(\x)\ge 0$ holds. Meanwhile, it is known that $(\x_u-\x_v)^2 \le 2(\x^2_u+\x^2_v)$ holds for each edge ${\langle u,v\rangle}\in \E$ and the equality holds when $\x_u=-\x_v$. Therefore, we have {\large$$\sup_{\x \in \R^n}\tfrac{\sum_{{\langle u,v\rangle}\in \E}(\x_u-\x_v)^2}{\sum_{u\in \V}\x^2_u d_u} \le 2,$$} \ie $f(\x) \le 1$ for any $\x \in \R^n$ , which completes the proof. 
\end{proof}

\subsection{Proof of Theorem~\ref{thm:frequencyratio}}
\begin{proof}
Given a connected graph $\G=(\V,\E)$ with a feature signal $\x$ and homophily ratio $h$, consider an optimal polynomial filter $\mathrm{F(\w)}$. Denote $\z=\mathrm{F(\w)}\x=\sum^K_{k=0}\w_k\P^k\x$. Thus we have {\large
$$f(\mathrm{F(\w)}\x) =f(\z) =\tfrac{\z^\top \L\z}{2}=\tfrac{\sum_{{\langle u,v\rangle}\in \E}(\z_u-\z_v)^2}{2\sum_{u\in \V}\z^2_u d_u}.$$} 
As $\mathrm{F(\w)}$ is the optimal filter for node classification, it is reasonable to assume that the filtered signals of nodes belonging to the same classes tend to cluster together, while signals of nodes from distinct classes exhibit a notable spatial separation.

W.l.o.g., for $\forall u,v \in \V$, we assume a constant $\delta$ such that $|\z_u - \z_v| \le c\delta$ with $c \ll 1$ if $\Y_u=\Y_v$; otherwise $|\z_u - \z_v|=g(\Y_u, \Y_v)\delta$ with $g(\Y_u, \Y_v)\ge 1$ where $\Y\in \mathbb{N}^{n\times |\C|}$ is the one-hot label matrix and $g(\cdot)$ is a function parameterized by $\Y_u$ and $\Y_v$. Consequently, the spectral frequency $f(\z)=\tfrac{\sum_{{\langle u,v\rangle}\in \E}(\z_u-\z_v)^2}{2\sum_{u\in \V}\z^2_u d_u}$ approaches to {\large$$\tfrac{c^2\delta^2 hm+\sum_{{\langle u,v\rangle}\in \E, \Y_u \neq \Y_v}g^2(\Y_u, \Y_v)\delta^2}{2\sum_{u\in \V}\z^2_u d_u}.$$} Since $c\ll 1$ and $g(\Y_u, \Y_v) \ge 1$, therefore $g^2(\Y_u, \Y_v) \gg c^2$ holds. In this regard, frequency $f(\textstyle\sum^K_{k=0}\w_k\P^k\x)$ is dominated by {\large$$\tfrac{\sum_{{\langle u,v\rangle}\in \E, \Y_u \neq \Y_v}g^2(\Y_u, \Y_v)\delta^2}{2\sum_{u\in \V}\z^2_u d_u},$$} \ie the number of edges connecting nodes from distinct classes, which is proportional to $1-h$.
\end{proof}

\subsection{Proof of Theorem~\ref{thm:homo}}
\begin{proof}
Consider a propagation matrix $\P$ and graph signal $\x$. Let $\{\lambda_1,\lambda_2,\cdots, \lambda_n\}$\footnote{For notation simplicity, we here reuse the notation $\lambda_i$ as the $i$-th eigenvalue of propagation matrix $\P$.} be the eigenvalues of $\P$ associated with eigenvectors $\{\v_1,\v_2,\cdots, \v_n\}$. For a general (non-bipartite) connected graph $\G$, we have $-1<\lambda_1 \le \lambda_2 \le \cdots \le \lambda_n =1$ and $\v^\top_i \v_j =0$ for $i \neq j$ and $\v^\top_i \v_j =1$ for $i = j$. In particular, we have $\lambda_n=1$ and $\v_n =\tfrac{\D^{\tfrac{1}{2}}\mathbf{1}}{\sqrt{2m}}$ where $\mathbf{1}\in \R^n$ is the all-one vector. Hence, $\P^k\x$ can be formulated as $\P^k\x=\sum^n_{i=1}\lambda^k_i(\v^\top_i\x)\v_i$. Specifically, we have \[\|\P^k\x\|=\sqrt{\left(\sum^n_{i=1}\lambda^k_i(\v^\top_i\x)\v_i \right)\cdot \left(\sum^n_{j=1}\lambda^k_j(\v^\top_j\x)\v_j \right)}=\sqrt{\sum^n_{i=1}\lambda^{2k}_i(\v^\top_i\x)^2}.\] Similarly, we have
\[\P^k\x \cdot \P^{k+1}\x =\left(\sum^n_{i=1}\lambda^k_i(\v^\top_i\x)\right)\cdot \left(\sum^n_{i=1}\lambda^{k+1}_i(\v^\top_i\x)\right)=\sum^n_{i=1}\lambda^{2k+1}_i(\v^\top_i\x)^2.\]  

Therefore, we have {\Large\[\tfrac{\P^k\x \cdot \P^{k+1}\x}{\|\P^k\x\| \|\P^{k+1}\x\|}=\tfrac{\sum^n_{i=1}\lambda^{2k+1}_i(\v^\top_i\x)^2}{\sqrt{\sum^n_{i=1}\lambda^{2k}_i(\v^\top_i\x)^2}\sqrt{\sum^n_{i=1}\lambda^{2k+2}_i(\v^\top_i\x)^2}}.\]} For ease of exposition, we denote $c_k=\|\P^k\x\|$ and let $t$ be the integer index such that $\lambda_{t}<0\le \lambda_{t+1}$. As a result, we have 
\begin{equation}\label{eqn:unitvector}
\sum^n_{i=1}\left(\tfrac{\lambda^k_i(\v^\top_i\x)}{c_k}\right)^2=\sum^n_{i=1}\left(\tfrac{\lambda^{k+1}_i(\v^\top_i\x)}{c_{k+1}}\right)^2=1    
\end{equation} and {\Large\[\tfrac{\P^k\x \cdot \P^{k+1}\x}{\|\P^k\x\| \|\P^{k+1}\x\|}=\sum^t_{i=1}\tfrac{\lambda^{2k+1}_i(\v^\top_i\x)^2}{c_kc_{k+1}}+\sum^n_{i=t}\tfrac{\lambda^{2k+1}_i(\v^\top_i\x)^2}{c_kc_{k+1}}.\]} Note that the exponent $2k+1$ remains odd integer for any $k$ values. W.l.o.g., we denote the negative part as function {\large$$f_N(t,k)=\sum^t_{i=1}\tfrac{\lambda^{2k+1}_i(\v^\top_i\x)^2}{c_kc_{k+1}}$$} and the positive part as function {\large$$f_P(t,k)=\sum^n_{i=t}\tfrac{\lambda^{2k+1}_i(\v^\top_i\x)^2}{c_kc_{k+1}}.$$}

Notice that $\lambda_i\in(-1,0)$ for $i\in \{1,\cdots,t\}$, $\lambda_i\in[0,1]$ ($\lambda_n=1$) for $i\in\{t+1,\cdots,n\}$, and $(\v^\top_i\x)^2$ for $i\in\{1,\cdots,n\}$ are constants. When $k$ increases, numerator $\sum^t_{i=1}\lambda^{2k}_i(\v^\top_i\x)^2$ of $\tfrac{\sum^t_{i=1} \lambda^{2k}_i(\v^\top_i\x)^2}{c^2_k}$ monotonically decreases and asymptotically approaches to $0$. According to Equation~\eqref{eqn:unitvector}, {\large$$\sum^t_{i=1}\tfrac{\lambda^{2k}_i(\v^\top_i\x)^2}{c^2_k}+\sum^n_{i=t}\tfrac{\lambda^{2k}_i(\v^\top_i\x)^2}{c^2_k}=1$$} holds for all $k$. Therefore, when the case $\sum^t_{i=1}\lambda^{2k}_i(\v^\top_i\x)^2 > \sum^n_{i=t}\lambda^{2k}_i(\v^\top_i\x)^2$ occurs for certain $k$, there exists an integer $\eta \in \mathbb{N}$ such that {\large$$\sum^t_{i=1}\lambda^{2(\eta-1)}_i(\v^\top_i\x)^2 > \sum^n_{i=t}\lambda^{2(\eta-1)}_i(\v^\top_i\x)^2$$} and {\large$$\sum^t_{i=1}\lambda^{2\eta}_i(\v^\top_i\x)^2 \le \sum^n_{i=t}\lambda^{2\eta}_i(\v^\top_i\x)^2.$$} Considering the monotonicity of both $\sum^t_{i=1}\lambda^{2k}_i(\v^\top_i\x)^2$ and $\sum^n_{i=t}\lambda^{2k}_i(\v^\top_i\x)^2$, when $k\ge \eta$, $f_N(t,k)$ monotonically decreases while $f_P(t,k)$ monotonically increases with $k$. As a consequence, $\tfrac{\P^k\x \cdot \P^{k+1}\x}{\|\P^k\x\| \|\P^{k+1}\x\|}$ is monotonically increasing with $k$, and thus the angle angle $\arccos\left(\tfrac{\P^k\x \cdot \P^{k+1}\x}{\|\P^k\x\| \|\P^{k+1}\x\|}\right)$ is progressively smaller. 

Meanwhile, when $K\to \infty$, we have {\large$$\lim_{K\to\infty}\P^K\x=\lambda^K_n (\v^\top_n\x) \v_n=\tfrac{\v^\top_n\x}{\sqrt{2m}}\D^{\tfrac{1}{2}}\mathbf{1}$$} where $\v_n =\tfrac{\D^{\tfrac{1}{2}}\mathbf{1}}{\sqrt{2m}}$. Therefore, $\lim_{K\to\infty}\arccos\left(\frac{\P^K\x \cdot \P^{K+1}\x}{\|\P^K\x\|\|\P^{K+1}\x\|}\right) \to 0 $ holds. 
\end{proof}

\subsection{Proof of Theorem~\ref{thm:pivot}}
\begin{proof}
We consider a $t$-regular graph $\G=(\V, \E)$ with $n$ nodes without self-loops. Given a random normalized signal $\x=(\x_1, \x_2, \cdots, \x_n)^\top$, $\phi \cdot \x =\sum^n_{i=1}\tfrac{\x_i}{\sqrt{n}}$ where $\phi \in \R^n$ is the normalized all-ones vector with $f(\phi)=0$. Meanwhile, recall that {\large$$f(\x)=\tfrac{\x^\top\L\x}{2}=\tfrac{\sum_{{\langle u,v\rangle}\in \E}(\x_u-\x_v)^2}{2\sum_{u\in \V}\x^2_u d_u}=\tfrac{\sum_{{\langle u,v\rangle}\in \E}(\x_u-\x_v)^2}{2t}$$} since $\sum_{u\in \V}\x^2_u=1$ and $d_u=t\ \forall u\in \V$. Over the randomness of $\G$, any two nodes $u,v\in \V$ is connected with probability $\tfrac{t}{n-1}$. Thus, the expectation of the spectral signal frequency is
\begin{align*}
\mathbb{E}_{\G\sim \mathcal{G}}[f(\x)]
& = \tfrac{1}{2t}\cdot \tfrac{t}{n-1}\sum_{u\in \V}\sum_{v\in \V\setminus\{u\}}\tfrac{(\x_u-\x_v)^2}{2} \\ \\
& = \tfrac{1}{4(n-1)}\sum_{u\in \V}\sum_{v\in \V\setminus\{u\}}(\x_u-\x_v)^2\\
& = \tfrac{1}{4(n-1)}\left(\sum_{u\in \V}(n-1)\x^2_u-\sum_{u\in \V}2\x_u\big(\sum_{v\in \V\setminus \{u\}}\x_v\big)\right) \\
\end{align*}
\begin{align*}
& = \tfrac{1}{4(n-1)}\left(n-1-\sum_{u\in \V}2\x_u\big(\sum^n_{i=1}\x_i-\x_u\big)\right)\\
& = \tfrac{1}{4(n-1)}\left(n-1-2\big(\sum^n_{i=1}\x_i\big)^2+2 \right)\\
& = \tfrac{1}{4(n-1)}\left(n+1-2\big(\sum^n_{i=1}\x_i\big)^2\right) \\
& = \tfrac{n+1}{4(n-1)}-\tfrac{1}{2(n-1)}\big(\sum^n_{i=1}\tfrac{\x_i}{\sqrt{n}}\big)^2 \\
& = \tfrac{n+1-2(\phi \cdot \x)^2}{4(n-1)}.
\end{align*}
As a consequence, if the angle $\theta:=\arccos( \phi \cdot \x )$ increases, $\phi\cdot \x$ decreases, resulting the increment of $\mathbb{E}_{\G\sim \mathcal{G}}[f(\x)]$, which completes the proof.
\end{proof}

\subsection{Proof of Theorem~\ref{thm:heteroproperty}}
Before the proof of Theorem~\ref{thm:heteroproperty}, we first introduce the following Lemma.
\begin{lemma}[Proposition 4.3~\citep{GuoW23}]~\label{lem:orthogonal} Vector $\v_k$ in Algorithm~\ref{alg:basis} is only dependent with $\v_{k-1}$ and $\v_{k-2}$.
\end{lemma}

Based on Lemma~\ref{lem:orthogonal}, it is intuitive that $\{\v_0, \v_1, \cdots,\v_K\}$ forms an orthonormal basis. Before proceeding, we establish the following lemma.
\begin{lemma}\label{lem:vkuk}
$\v_{k+1}$ constructed in Algorithm~\ref{alg:basis} is orthogonal to basis set $\{\u_0,\u_1, \cdots, \u_k\}$ for $k\in \{0,1,\cdots, K\!-\!1\}$.     
\end{lemma}
\begin{proof}[Proof of Lemma~\ref{lem:vkuk}]
In particular, we have $\u_0 = \v_0 = \tfrac{\x}{\|\x\|}$ and {\large\[\u_k=\tfrac{\s_{k-1}/k+t_k\v_k}{\|\s_{k-1}/k+t_k\v_k\|}=\tfrac{\frac{1}{k}\sum^{k-1}_{i=0}\u_i+t_k\v_k}{\|\frac{1}{k}\sum^{k-1}_{i=0}\u_i+t_k\v_k\|}\]} for $k\in [K]$ according to Algorithm~\ref{alg:basis}. By recursively replacing $\u_i$ with $\{\v_0, \v_1, \cdots,\v_i\}$, we obtain that $\u_k$ can be constructed by $\v_i$. In this regard, let $\{\alpha_0, \alpha_1, \cdots, \alpha_k\} \in \R$ be constants such that $\u_k=\sum^k_{i=0}\alpha_i \v_i$ holds. Since $\{\v_0, \v_1, \cdots,\v_K\}$ is an orthonormal basis, therefore $\v_{k+1}$ is orthogonal to $\{\u_0,\u_1, \cdots, \u_k\}$ as $\v_{k+1} \cdot \u_k =\v_{k+1} \cdot (\sum^k_{i=0}\alpha_i \v_i) =0$.    
\end{proof}

\begin{proof}[Proof of Theorem~\ref{thm:heteroproperty}]
Let $\theta:=\tfrac{(1-\hat{h})\pi}{2}$ be the predefined angle. First, we prove that $\u_0 \cdot \u_1 = \cos\theta$. In particular, we have $\u_1=\tfrac{\u_0+t_1\v_1}{\|\u_0+t_1\v_1\|}$ and $t_1=\sqrt{(\tfrac{\s_0^\top \u_0}{\cos\theta})^2-1}=\sqrt{\tfrac{1}{\cos^2\theta}-1}$. Then 
{\large\[\|\u_0+t_1\v_1\|=\sqrt{\u_0\cdot\u_0+t^2_1\v_1\cdot \v_1+2t_1\u_0 \cdot \v_1}=
\sqrt{1+t^2_1}=\tfrac{1}{\cos\theta}\]} as $\u_0 \cdot \v_1=0$ according to Lemma~\ref{lem:vkuk}.  Hence, $\u_0 \cdot \u_1 = \u^\top_0(\u_0+t_1\v_1)\cos\theta=\cos\theta$. 

Second, we assume that $\u_i \cdot \u_j=\cos\theta$ holds for $\forall i,j\in\{0,1,\cdots, k\!-\!1\}$ and $i\neq j$. In what follows, we then prove that $\u_k \cdot \u_j=\cos\theta$ holds for $j\in\{0,1,\cdots, k\!-\!1\}$. Specifically, we have {\large$$\u_k=\tfrac{\frac{1}{k}\sum^{k-1}_{i=0}\u_i+t_k\v_k}{\|\frac{1}{k}\sum^{k-1}_{i=0}\u_i+t_k\v_k\|}$$} where \[t_k=\sqrt{\left(\tfrac{\s_{k-1}^\top\u_{k-1}}{k\cos\theta}\right)^2-\tfrac{(k-1)\cos\theta+1}{k}}, \textrm{ and }\ \s_{k-1}=\sum^{k-1}_{i=0}\u_i.\] In particular, for the denominator of $\u_k$, we have
\begin{align*}
\left\|\frac{1}{k}\sum^{k-1}_{i=0}\u_i+t_k\v_k\right\| 
& = \sqrt{\left(\frac{1}{k}\sum^{k-1}_{i=0}\u^\top_i+t_k\v^\top_k\right)\left(\frac{1}{k}\sum^{k-1}_{i=0}\u_i+t_k\v_k\right)} \\
& = \sqrt{\tfrac{\sum^{k-1}_{i=0}\u^\top_i \u_i+2\sum^{k-2}_{i=0}\u^\top_i(\sum^{k-1}_{j=i+1}\u_j)}{k^2}+t^2_k} \\
& = \sqrt{\tfrac{k+k(k-1)\cos\theta}{k^2}+\left(\tfrac{\s_{k-1}^\top\u_{k-1}}{k\cos\theta}\right)^2-\tfrac{(k-1)\cos\theta+1}{k}} \\
& = \frac{\sum^{k-1}_{i=0}\u^\top_i \cdot \u_{k-1} }{k\cos\theta} \\
& = \tfrac{1+(k-1)\cos\theta}{k\cos\theta}.
\end{align*}
Meanwhile, we have {\large\[\u_k \cdot \u_j  = \tfrac{\frac{1}{k}\sum^{k-1}_{i=0}\u^\top_i \u_j+t_k\v^\top_k \u_j}{\|\frac{1}{k}\sum^{k-1}_{i=0}\u_i+t_k\v_k\|} = \tfrac{\frac{1}{k} (1+(k-1)\cos\theta)\cdot k\cos\theta}{1+(k-1)\cos\theta}= \cos\theta.\]} 

Eventually, it is easy to verify that the derivation holds for $\forall j\in\{0,1,\cdots, k\!-\!1\}$, which completes the proof.
\end{proof}

\subsection{Proofs of Over-smoothing and Over-Squashing}
\begin{proof}[Proof of Theorem~\ref{thm:oversmoothing}]
Given a feature matrix $\X^k$ from the $k$-th layer, over-smoothing occurs if Dirichlet energy $E(\G,\X^k)\to 0$ for a sufficiently large $k$. In what follows, we prove that $E(\G,\X^k)$ of $\X^k$ obtained by \ours remains constant when $k\to \infty$. Specifically, let $\X^k=(\tau\P^k+ (1-\tau)\P_{\theta,k})\X$ be the $k$-th layer propagated feature matrix where $\P$ is the propagation matrix and $\P_{\theta,k}$ is the $k$-th rotation matrix. 

For $E(\G,\X^k)$ and $k\to \infty$, we have 
\begin{align*}
E(\G,\X^k)
& = \tfrac{1}{n}\sum_{v\in \V}\sum_{u\in \N_v}\|\X^{(k)}_v-\X^{(k)}_u\|^2_2 \\
& = \tfrac{1}{n}\sum_{v\in \V}\sum_{u\in \N_v}\|\tau\P^k\X[v]-\tau\P^k\X[u]+(1-\tau)\P_{\theta,k}\X[v]-(1-\tau)\P_{\theta,k}\X[u]\|^2_2. 
\end{align*}
When $k\to \infty$, $\lim_{K\to\infty}\P^K\x$ converges to $\v_n$, as pointed out in the proof of Theorem~\ref{thm:homo}. Therefore $\tau\P^k\X[v]=\tau\P^k\X[u]$. Thus we have
\begin{align*}
\lim_{K\to\infty}E(\G,\X^k)
& =\tfrac{1}{n}\sum_{v\in \V}\sum_{u\in \N_v}\|(1-\tau)\P_{\theta,k}\X[v]-(1-\tau)\P_{\theta,k}\X[u]\|^2_2    \\
& = \tfrac{1-\tau}{n}\sum_{v\in \V}\sum_{u\in \N_v}\|\P_{\theta,k}\X[v]-\P_{\theta,k}\X[u]\|^2_2
\end{align*}


According to Definition~\ref{def:rotation}, there exists a unitary matrix $\U\in\mathbf{C}^{n\times n}$ such that $\mathbf{R}(\theta):=\U^{-1}\P_{\theta,k}\U$. Therefore, we have {\large$$\|\P_{\theta,k}\X[v]-\P_{\theta,k}\X[u]\|^2_2 = \|\U\mathbf{R}(\theta)\U^{-1}\X[v]-\U\mathbf{R}(\theta)\U^{-1}\X[u]\|^2_2.$$} Notice that both $\mathbf{R}(\theta)$ and $\U^{-1}$ are unitary matrices. According to the isometry property of the unitary matrix, we have 
\begin{align*}
\lim_{K\to\infty}E(\G,\X^k)
& = \tfrac{1-\tau}{n}\sum_{v\in \V}\sum_{u\in \N_v}\|\U\mathbf{R}(\theta)\U^{-1}\X[v]-\U\mathbf{R}(\theta)\U^{-1}\X[u]\|^2_2 \\
& = \tfrac{1-\tau}{n}\sum_{v\in \V}\sum_{u\in \N_v}\|\mathbf{R}(\theta)\U^{-1}\X[v]-\mathbf{R}(\theta)\U^{-1}\X[u]\|^2_2 \\
& = \tfrac{1-\tau}{n}\sum_{v\in \V}\sum_{u\in \N_v}\|\U^{-1}\X[v]-\U^{-1}\X[u]\|^2_2 \\
& = \tfrac{1-\tau}{n}\sum_{v\in \V}\sum_{u\in \N_v}\|\X_v-\X_u\|^2_2
\end{align*}
As a consequence, $\lim_{K\to\infty}E(\G,\X^k)=(1-\tau)E(\G,\X)$, which completes the proof.
\end{proof}

\begin{proof}[Proof of Theorem~\ref{thm:oversquashing}]
As discussed, Jacobian $|\partial \z^{(k)}_u / \partial \x_v|$ measures the sensitivity of information received at node $u$ to the signal of node $v$ when propagating $k$ distance. Over-squashing occurs if $|\partial \z^{(k)}_u / \partial \x_v| \le c \cdot (\P^k_M)_{uv}$ where $c$ is a constant and $\P_M$ is the message-passing matrix, \ie the sensitivity (node dependence) decays exponentially to propagation distance. According to Equation~\eqref{eqn:convoform}, we have {\large$$\left|\tfrac{\partial \z^{(k)}_u}{\partial \x_v}\right|=\w_k\big(\tau(\P^k)_{uv}+ (1-\tau)(\P_{\theta,k})_{uv}\big)$$} in \ours. Here, $\w_k$ as the learnable parameter can be seen as a constant. Note that the rotation matrix $\P_{\theta,k}$ is determined by rotation angle $\theta$ and rotation axis without utilizing the underlying graph topology. Therefore, the value of $(\P_{\theta,k})_{uv}$ is independent of propagation step $k$, which completes the proof.
\end{proof}

\clearpage

\section{Experimental Settings}\label{app:settings}

\begin{table*}[!t]
\centering
\caption{Dataset details.} \label{tbl:dataset}\vspace{-1mm} 
\setlength{\tabcolsep}{0.5em}
\small
\resizebox{0.98\textwidth}{!}{%
\begin{tabular} {@{}l|rrrrrrrrr@{}}
\toprule
{\bf Dataset}  & \multicolumn{1}{c}{Cora} & \multicolumn{1}{c}{Citeseer} & \multicolumn{1}{c} {Pubmed} &\multicolumn{1}{c}{Actor}& \multicolumn{1}{c} {Chameleon}& \multicolumn{1}{c}{Squirrel}& \multicolumn{1}{c}{Penn94} & \multicolumn{1}{c}{Genius} &  \multicolumn{1}{c}{Ogbn-arxiv}\\ \midrule 
{\bf{\#Nodes ($\boldsymbol{n}$)}}  & 2,708 & 3,327& 19,717 & 7,600 & 2,277 & 5,201 & 41,554 & 421,961 & 169,343 \\
{\bf{\#Edges ($\boldsymbol{m}$)}} &  5,429 &  4,732& 44,338 & 26,659 & 31,371 & 198,353 & 1,362,229 & 984,979 & 1,166,243 \\
{\bf \#Features ($\boldsymbol{d}$)} & 	1,433 & 3,703& 500 & 932 & 2,325 & 2,089  & 5 & 12 & 128  \\
{\bf \#Classes} & 7 & 6& 3& 5& 5& 5 & 2 & 2 & 40 \\
{\bf Homo. ratio ($\boldsymbol{h}$)}& 0.81 &  0.74&  0.80 &  0.22 &  0.23 & 0.22  & 0.47  & 0.62 & 0.65 \\ \bottomrule
\end{tabular}}
\end{table*}

\spara{Datasets} Table~\ref{tbl:dataset} presents the detailed statistics of the $9$ real-world datasets tested in our experiments. The three homophily datasets, \ie Cora, Citeseer, and Pubmed are citation networks with homophily ratios of $0.81$, $0.74$, and $0.80$ respectively. Each graph node represents a research paper, and each edge denotes a citation relationship. Feature vectors of nodes are bag-of-words representations. The one-hot label assigned to each node stands for one research field of the paper. The rest three small datasets \ie Actor, Chameleon, and Squirrel are heterophily datasets with homophily ratios of $0.22$, $0.23$, and $0.22$ respectively. Specifically, Actor is a co-occurrence graph from the film-director-actor-writer network from WebKB3~\citep{TangSWY09,PeiWCLY20}. Squirrel and Chameleon are two datasets extracted from Wikipedia web pages, and nodes are categorized by the average amounts of monthly traffic~\citep{LiZCSLLQ22}. The additional three large datasets, \ie Penn94, Genius, and Ogbn-arxiv are with homophily ratios of $0.47$, $0.62$, and $0.65$ respectively. Specifically, Penn94 and Genius are from the LINKX datasets~\cite{lim2021large}, and Ogbn-arxiv is a citation network from~\cite{hu2020open}.  

\spara{Running environment} All our experiments are conducted on a Linux machine with an NVIDIA RTX A5000 (24GB memory), Intel Xeon(R) CPU (2.80GHz), and 500GB RAM.

\spara{Parameter settings} During the training process, learnable parameters are tuned with Adam~\citep{KingmaB14} optimizer. We set a patience of early stopping with $200$ epochs. For hyperparameters, we fix propagation hop $K=10$ for all tested datasets. The rest hyperparameters are selected in the following ranges. 

1. Learning rate: $[0.001, 0.005, 0.01, 0.05, 0.1, 0.15, 0.2]$;

2. Hidden dimension: $[64, 128, 256]$;

3. MLP layers: $[2,3,4,5,6]$;

4. Weight decays: $[0, 1e-4, 5e-4, 0.001]$;

5. Drop rates: $[0,0.1,0.2,\cdots, 0.9]$.

\spara{Choice of $\tau$} Ideally, the selection of $\tau$ for each dataset is highly related to its homophily ratio. In our experiments, the $\tau$ values are set as in Table~\ref{tbl:tauselection}.

\spara{Homophily ratio estimation} For each dataset, $10$ random splits of training/validation/test are generated. The homophily ratio is estimated using the training set from each split independently as the input to \ours for the construction of polynomial bases. The reported accuracy score is average with a standard deviation over the results on the $10$ splits of each tested dataset.

\begin{table}[!t]
\centering
 \caption{Selections of $\tau$ for the tested datasets.} \label{tbl:tauselection}\vspace{-1mm} 
    \setlength{\tabcolsep}{0.5em}
 \small
\begin{tabular} {@{}c|cccccccccc@{}}
\toprule
{\bf Dataset}  & \multicolumn{1}{c}{Cora} & \multicolumn{1}{c}{Citeseer} & \multicolumn{1}{c} {Pubmed}  &\multicolumn{1}{c}{Actor}& \multicolumn{1}{c} {Chameleon}& \multicolumn{1}{c}{Squirrel} &\multicolumn{1}{c}{Penn94}& \multicolumn{1}{c} {Genius}& \multicolumn{1}{c}{Ogbn-arxiv}\\ \midrule 
{\bf{$\tau$}}  & 1.0 & 0.9 & 0.8 & 0.1 & 0.7 & 0.7 & 0.9 & 0.0 & 0.5\\ \bottomrule
\end{tabular}
\end{table}

\section{Additional Experiments}\label{app:exp}

\begin{table*}[!t]
\centering
 \caption{Accuracy (\%) on large datasets.}\label{tbl:largedata}\vspace{-2mm} 
 \normalsize
  \renewcommand\arraystretch{1.3}
\begin{tabular}{@{}c|c|c|c@{}}
\toprule
\multicolumn{1}{c|}{Methods} & \multicolumn{1}{c}{Penn94} & \multicolumn{1}{c}{Genius}&  \multicolumn{1}{c}{Ogbn-arxiv} \\ \midrule
\GCN          & 82.47 $\pm$ 0.27 &  87.42 $\pm$ 0.37 &	71.74 $\pm$ 0.29     \\
\SGC          &  81.44 $\pm$ 0.15 &   87.59	$\pm$ 0.10 & 71.21 $\pm$ 0.17   \\ 
\SIGN         & 82.13 $\pm$ 0.28  & 88.19 $\pm$ 0.02&	   71.95 $\pm$ 0.12       	\\
\ChebNet      & 82.59 $\pm$ 0.31 &  89.36 $\pm$ 0.31 &	   71.12 $\pm$ 0.22      \\
\GPRGNN       & 83.54 $\pm$ 0.32 & 90.15 $\pm$ 0.30 &	   71.78 $\pm$ 0.18     	 \\
\BernNet      & 83.26 $\pm$ 0.29 & 90.47 $\pm$ 0.33 &	   71.96 $\pm$ 0.27    	 \\
\ChebNetII    &  \bf{84.86 $\pm$ 0.33} & \underline{90.85 $\pm$ 0.32 }  &\underline{72.32 $\pm$ 0.23}\\
\OptBasisGNN  &  \underline{84.85 $\pm$ 0.39} & 90.83 $\pm$ 0.11  &	   72.27 $\pm$ 0.15 \\
\ours         & 84.46 $\pm$ 0.33 & \bf{90.94 $\pm$ 0.22}&	   \bf{73.70 $\pm$ 0.70}     \\ \bottomrule 
\end{tabular}

\footnotemark{The highest score is in bold while the second-highest score is underlined for each dataset.}
\end{table*}

\spara{Comparison on large datasets} For a comprehensive evaluation, we add three large datasets, \ie Penn94, Genius, and Ogbn-arxiv which are summarized in Table~\ref{tbl:dataset}. For large graphs,  we stick to tuning parameters by following the parameter settings in Appendix~\ref{app:settings}. In particular, we compare \ours against existing polynomial filters. The results are presented in Table~\ref{tbl:largedata}. As demonstrated, \ours achieves the highest scores on Genius and Obgn-arxiv and approaches the highest score on Penn94. Specifically, \ours advances the accuracy score up to $1.38\%$ on Ogbn-arxiv. Those observations further confirm the universality and efficacy of \newbasis.

\spara{Robustness of homophily ratio estimation} To further verify the robustness of the estimation, we vary the percentages of the training set in $\{10\%, 20\%, 30\%, 40\%, 50\%, 60\%\}$ on Cora and Squirrel respectively. Next, we generate $10$ random training sets in each of the above percentages. Subsequently, we average the estimated homophily ratios $\hat{h}$ over the $10$ splits and report the results in Table~\ref{tbl:VaryEstHomRatio}. As shown, the estimated homophily ratio $\hat{h}$ is approaching the true homophily ratio $h$ across varying percentages of training data. This observation verifies that a high-quality estimation of the homophily ratio is accessible effectively by a small proportion of training sets.

\begin{table}[!t]
\centering
 \caption{Estimated homophily ratio $\hat{h}$ over varying training percentages on Cora and Squirrel.} \label{tbl:VaryEstHomRatio}\vspace{-1mm} 
    \setlength{\tabcolsep}{0.5em}
 \small
 \renewcommand\arraystretch{1.5}
\begin{tabular} {@{}l|rrrrrr@{}}
\toprule
{\bf Dataset}  & \multicolumn{1}{c}{$10\%$} & \multicolumn{1}{c}{$20\%$} & \multicolumn{1}{c} {$30\%$} &\multicolumn{1}{c}{$40\%$}& \multicolumn{1}{c} {$50\%$}& \multicolumn{1}{c}{$60\%$}\\ \midrule 
{Cora}  & 0.83 $\pm$ 0.05 & 0.83 $\pm$0.04 & 0.83 $\pm$ 0.03 & 0.83 $\pm$ 0.01 & 0.82 $\pm$ 0.08 & 0.82$\pm$ 0.01 \\ \hline
{Squirrel} & 0.23 $\pm$  0.014 & 0.22 $\pm$  0.011 & 0.22 $\pm$  0.010 & 0.22 $\pm$  0.006 & 0.22 $\pm$  0.005 & 0.22 $\pm$ 0.005 \\ \bottomrule
\end{tabular}
\end{table}

\begin{table}[!t]
\centering
 \caption{Estimated homophily ratio $\hat{h}$ over varying training percentages on Cora and Squirrel.} \label{tbl:SenseRatio}\vspace{-1mm} 
    \setlength{\tabcolsep}{0.5em}
 \small
 \renewcommand\arraystretch{1.5}
\begin{tabular} {@{}l|rrrrr@{}}
\toprule
{\bf Varying $\hat{h}$}  & \multicolumn{1}{c}{$0.78$} & \multicolumn{1}{c}{$0.79$} & \multicolumn{1}{c} {$0.80$} &\multicolumn{1}{c}{$0.81$}& \multicolumn{1}{c} {$0.82$}\\ \midrule 
{Squirrel}  & 91.03 $\pm$ 0.61 & 91.28 $\pm$ 0.64 & 91.34 $\pm$ 0.62 & 91.19 $\pm$ 0.67 & 91.17 $\pm$ 0.69 \\ \hline
{\bf Varying $\hat{h}$}  & \multicolumn{1}{c}{$0.19$} & \multicolumn{1}{c}{$0.20$} & \multicolumn{1}{c} {$0.21$} &\multicolumn{1}{c}{$0.22$}& \multicolumn{1}{c} {$0.23$}\\ \midrule 
{Pubmed}  & 66.22 $\pm$ 1.43 & 66.96 $\pm$ 1.38 & 67.09 $\pm$ 1.08 & 67.01 $\pm$ 1.25 & 66.69 $\pm$ 1.26 \\ \bottomrule
\end{tabular}
\end{table}

\spara{Sensitivity of \ours to $\hat{h}$} We conduct an ablation study on the sensitivity of our proposed \ours to the estimated homophily ratio. Since \ours only utilizes the homophily basis on Cora ($\tau=1$ for Cora as shown in Table~\ref{tbl:tauselection}) and does not rely on the homophily ratio, we thus test \ours on Pubmed and Squirrel instead. To this end, we vary the estimated homophily ratio $\hat{h} \in \{0.78, 0.79, 0.80, 0.81, 0.82\}$ on Pubmed and  $\hat{h} \in\{0.19,0.20,0.21,0.22,0.23\}$ on Squirrel according to the estimation variances in Table~\ref{tbl:VaryEstHomRatio}. The achieved accuracy scores are presented in Table~\ref{tbl:SenseRatio}.

In Table~\ref{tbl:SenseRatio}, \ours demonstrates consistent accuracy scores across different homophily ratios for both Pubmed and Squirrel. Notably, the accuracy variations remain minor, staying within a $1\%$ range from the scores under the true homophily ratio, particularly on the Pubmed dataset.

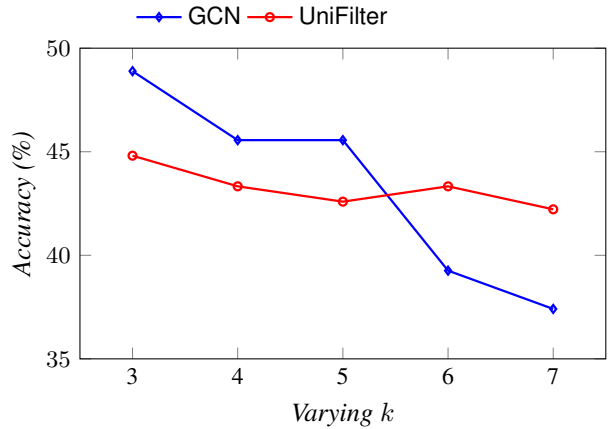
\begin{figure}[!t]
    \centering
   \begin{tikzpicture}[scale=1,every mark/.append style={mark size=1.5pt}]
    \begin{axis}[
        height=\columnwidth/3,
        width=\columnwidth/2,
        ylabel={\em Accuracy (\%)},
        xlabel={\em Varying $k$},
        xmin=2.5, xmax=7.5,
        ymin=35, ymax=50,
        xtick={3,4,5,6,7},
        xticklabel style = {font=\small},
        yticklabel style = {font=\small},
        xticklabels={ 3,4,5,6,7},
        ylabel style={yshift=-0.1cm},
        legend columns=2,
        legend style={fill=none,font=\footnotesize,at={(0.35,1.1)},anchor=center,draw=none},
    ] 
           \addplot[line width=0.3mm,mark=diamond,color=blue]
        coordinates {
            (3, 48.89) (4, 45.56) (5,45.56) (6, 39.26) (7, 37.41)}; 
        \addplot[line width=0.3mm,mark=o,color=red]
        coordinates {
            (3, 44.81) (4, 43.33) (5,42.59) (6, 43.33) (7, 42.22)}; 
    \legend{\textsf{\GCN}, \textsf{\ours}}
    \end{axis}
\end{tikzpicture}
\caption{Accuracy (\%) with varying $k$.} \label{fig:oversquashing}
\end{figure}

\spara{Mitigation of Over-squashing} To validate that \ours is able to mitigate the over-squashing issue, we customize a synthetic graph for node classification.  Specifically, we generate a binary tree of depth $7$ empirically as the testing graph, denoted as $\G_B$. In particular, $\G_B$ consists of $127$ nodes. Each node is assigned a random $100$-dimension one-hot vector as a node feature and a random label from the label set $\{0,1,2\}$. We vary the propagation hop $k\in \{3,4,5,6,7\}$. As expected, the number of neighbors exponentially increases with the increment of $k$. In this ablation study, we compare \ours with \GCN and tune their hyperparameters for the best possible performance on $\G_B$. The accuracy scores are plotted in Figure~\ref{fig:oversquashing}. As illustrated, the accuracy scores of \GCN decrease along the increase of $k$, especially from $k=5$ to $k=6$. In contrast, \ours demonstrates consistency performance across the varying $k$. This observation aligns with the conclusion from Theorem~\ref{thm:oversquashing} which posits that the Jacobian of the propagation matrix within \ours is independent of propagation step $k$.

\end{sloppy}
\end{document}